\pdfoutput=1

\documentclass[11pt]{article}

\usepackage{acl}

\usepackage{times}
\usepackage{latexsym}

\usepackage[T1]{fontenc}

\usepackage[utf8]{inputenc}

\usepackage{microtype}

\usepackage{amsmath}
\usepackage{amssymb}
\usepackage{graphicx}
\usepackage{gensymb}
\usepackage{transparent}
\usepackage{subcaption}
\usepackage{booktabs}
\usepackage{CJKutf8}
\usepackage[utf8]{inputenc}
\usepackage[main=english]{babel}
\usepackage{multirow}
\usepackage{pdfcomment}
\usepackage{todonotes}
\usepackage[ruled,vlined,linesnumbered]{algorithm2e}

\newcommand{\bap}{\textsc{BAP}}
\newcommand{\dirprobe}{\textsc{DirProbe}}
\newcommand{\depprobe}{\textsc{DepProbe}}
\DeclareMathOperator*{\argmax}{argmax\,}
\DeclareMathOperator*{\argmin}{argmin\,}

\title{Probing for Labeled Dependency Trees}

\author{Max M{\"u}ller-Eberstein  \and Rob van der Goot \and Barbara Plank \\
  Department of Computer Science \\
  IT University of Copenhagen, Denmark \\
  \texttt{mamy@itu.dk, robv@itu.dk, bapl@itu.dk} \\}

\begin{document}
\maketitle
\begin{abstract}
Probing has become an important tool for analyzing representations in Natural Language Processing (NLP). For graphical NLP tasks such as dependency parsing, linear probes are currently limited to extracting undirected or unlabeled parse trees which do not capture the full task. This work introduces \depprobe{}, a linear probe which can extract \textit{labeled} and \textit{directed} dependency parse trees from embeddings while using fewer parameters and compute than prior methods. Leveraging its full task coverage and lightweight parametrization, we investigate its predictive power for selecting the best transfer language for training a full biaffine attention parser. Across 13 languages, our proposed method identifies the best source treebank 94\% of the time, outperforming competitive baselines and prior work. Finally, we analyze the informativeness of task-specific subspaces in contextual embeddings as well as which benefits a full parser's non-linear parametrization provides.
\end{abstract}

\section{Introduction}

Pre-trained, contextualized embeddings have been found to encapsulate information relevant to various syntactic and semantic tasks out-of-the-box \citep{tenney2019,hewitt2019}. Quantifying this latent information has become the task of \textit{probes} --- models which take frozen embeddings as input and are parametrized as lightly as possible (e.g.\ linear transformations). Recent proposals for edge probing \citep{tenney2019} and structural probing \citep{hewitt2019} have enabled analyses beyond classification tasks, including graphical tasks such as dependency parsing. They are able to extract dependency graphs from embeddings, however these are either undirected \citep{hewitt2019,maudslay2020} or unlabeled \citep{kulmizev-etal-2020-neural}, thereby capturing only a subset of the full task.

\begin{figure}
    \centering
    \pdftooltip{\includegraphics[width=.47\textwidth]{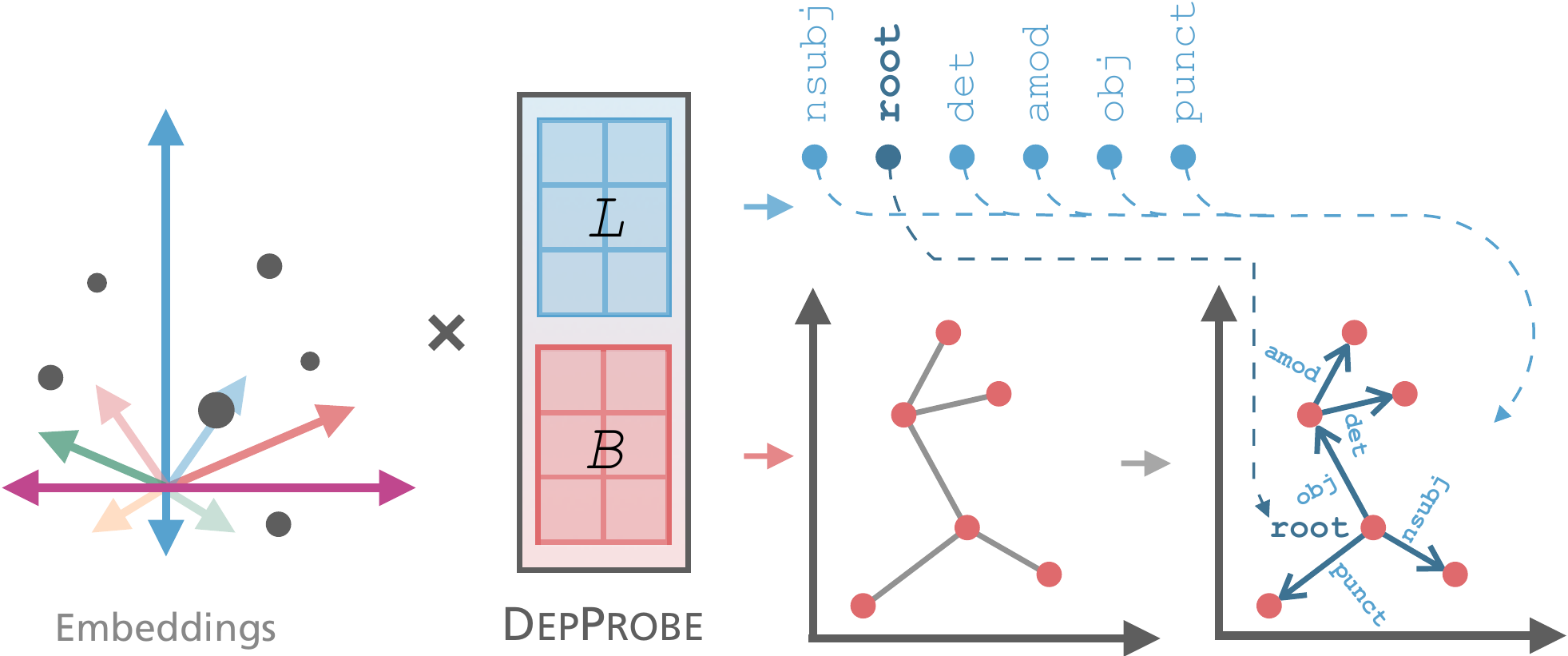}}{Screenreader Caption: Embeddings are multiplied with matrix B and L to transform them into a tree structural subspace and label subspace. Node with highest root probability sets directionality of all edges by pointing them away. Edges are labeled according to child embedding label from L.}
    \caption{\textbf{\depprobe{}} extracts tree structure using transformation $B$, labels using $L$ and infers directionality using \texttt{root}, based on contextualized embeddings.}
    \label{fig:depprobe}
\end{figure}

In this work, we investigate whether this gap can be filled and ask: \textit{Can we construct a lightweight probe which can produce fully directed and labeled dependency trees?} Using these trees, we further aim to study the less examined problem of transferability estimation for graphical tasks, extending recent work targeting classification and regression tasks \citep{nguyen2020,you2021}. Specifically: \textit{How well do our probe's predictions correlate with the transfer performance of a full parser across a diverse set of languages?}

To answer these questions, we contribute \depprobe{} (Figure \ref{fig:depprobe}), the first linear probe to extract directed and labeled dependency trees while using fewer parameters than prior work and three orders of magnitude fewer trainable parameters than a full parser (Section \ref{sec:probing}). As this allows us to measure labeled attachment scores (LAS), we investigate the degree to which our probe is predictive of cross-lingual transfer performance of a full parser across 13 typologically diverse languages, finding that our approach chooses the best transfer language 94\% of the time, outperforming competitive baselines and prior work (Section \ref{sec:experiments}). Finally, we perform an in-depth analysis of which latent information is most relevant for dependency parsing as well as which edges and relations benefit most from the expressivity of the full parser (Section \ref{sec:analysis}).\footnote{Code available at \href{https://personads.me/x/acl-2022-code}{https://personads.me/x/acl-2022-code}.}

\section{Related Work}

Given the ubiquitous use of contextualized embeddings \citep{devlin-etal-2019-bert,conneau-etal-2020-unsupervised,xue-etal-2021-mt5}, practitioners have turned to various methods for analyzing their linguistic features \citep{rogers-etal-2020-primer}. \citet{hewitt2019} examine these intrinsic properties in greater detail for English dependency parsing using a \textit{structural probe}, finding that \textit{undirected} dependency graphs are recoverable from BERT by learning a linear transformation on its embeddings (Section \ref{sec:undirected-probing}).

Extending the structural probe of \citet{hewitt2019} to 12 languages, \citet{chi2020} extract \textit{undirected} dependency graphs from mBERT \citep{devlin-etal-2019-bert}, further showing that head-to-child difference vectors in the learned subspace cluster into relations from the Universal Dependencies taxonomy \citep{de-marneffe-etal-2014-universal}.

Building on both the structural and tree depth probes \citep{hewitt2019}, \citet{kulmizev-etal-2020-neural} extract \textit{directed} dependency graphs from mBERT for 13 languages (Section \ref{sec:directed-probing}). Further variations to structural probing include regularization of the linear transformation \citep{limisiewicz2021} as well as alternative objective functions \citep{maudslay2020}.

None of the proposed linear probing approaches so far are able to produce full dependency parse trees (i.e.\ directed and labeled), however the closer a probe approximates the full task, the better it quantifies relevant information \citep{maudslay2020}. It would for example be desirable to estimate LAS for parsing a target treebank with a model trained on a different source without having to train a resource-intensive parser (e.g.\ \citealp{dozat2017}) on each source candidate. Although performance prediction methods for such scenarios exist, they typically do not cover graph prediction \citep{nguyen2020,you2021}.

In order to bridge the gap between full parsers and unlabeled probes, in addition to the gap between full fine-tuning and lightweight performance prediction, this work proposes a linear probe which can extract \textit{labeled} and \textit{directed} dependency parse trees while using less compute than prior methods (Section \ref{sec:probing}). We use our probe's LAS to evaluate its predictive power for full parser performance and leverage its linear nature to investigate how dependencies are represented in subspaces of contextual embeddings (Section \ref{sec:analysis}).

\section{Probing for Dependencies}\label{sec:probing}

In order to construct a directed and labeled dependency parse tree for a sentence $s$ consisting of the words $\{w_0, \dots, w_N\}$, we require information on the presence or absence of edges between words, the directionality of these edges $(\overrightarrow{w_i, w_j})$, and the relationships $\{r_0, \dots, r_N\}$ which they represent. Using the contextualized embeddings $\{\boldsymbol{h}_0, \dots, \boldsymbol{h}_N\}$ with $\boldsymbol{h}_i \in \mathbb{R}^{e}$, prior probing work has focused on the first step of identifying edges (Section \ref{sec:undirected-probing}) and later directionality (Section \ref{sec:directed-probing}). In this work, we propose a probe which completes the final relational step (Section \ref{sec:relational-probing}) and simultaneously provides a more efficient method for identifying directionality (Section \ref{sec:depprobe-method}).

\subsection{Undirected Probing}\label{sec:undirected-probing}

The structural probe introduced by \citet{hewitt2019} recovers the first piece of information (i.e.\ the undirected graph) remarkably well. Here, the probe is a linear transformation $B \in \mathbb{R}^{e \times b}$ with $b < e$ which maps contextual embeddings into a subspace in which the distance measure

\begin{equation}\label{eq:distance}
    \pdftooltip{d_B(\boldsymbol{h}_i, \boldsymbol{h}_j) = \sqrt{(B\boldsymbol{h}_i - B\boldsymbol{h}_j)^T (B\boldsymbol{h}_i - B\boldsymbol{h}_j)}}{Screenreader Caption: Equation: d subscript B of h subscript i and h subscript j equals square-root of B times h subscript i - B times h subscript j transposed times B times h subscript i - B times h subscript j.}
\end{equation}

between $\boldsymbol{h}_i$ and $\boldsymbol{h}_j$ is optimized towards the distance between two words in the dependency graph $d_P(w_i, w_j)$, i.e.\ the number of edges between the words. For each sentence, the loss is defined as the mean absolute difference across all word pairs:

\begin{equation}\label{eq:distance-loss}
    \pdftooltip{\mathcal{L}_B(s) = \frac{1}{N^2} \sum_{i=0}^N \sum_{j=0}^N \big| d_P(w_i, w_j) - d_B(\boldsymbol{h}_i, \boldsymbol{h}_j) \big|\hspace{.1em}\text{.}}{Screenreader Caption: Equation: L subscript B of s equals 1 over N squared times sum from indices i and j equal 0 to N of absolute value of d subscript P of w subscript i and w subscript j - d subscript B of h subscript i and h subscript j.}
\end{equation}

In order to extract an undirected dependency graph, one computes the distances for a sentence's word pairs using $d_B$ and extracts the minimum spanning tree (\citealp{jarnik1930,prim1957}; MST).

\subsection{Directed Probing}\label{sec:directed-probing}

Apart from the structural probe $B$, \citet{hewitt2019} also probe for tree depth. Using another matrix $C \in \mathbb{R}^{e \times c}$, a subspace is learned in which the squared $L_2$ norm of a transformed embedding $\lVert C\boldsymbol{h}_i\rVert^2_2$ corresponds to a word's depth in the tree, i.e.\ the number of edges from the root.

\citet{kulmizev-etal-2020-neural} combine the structural and tree depth probe to extract directed graphs. This directed probe (\dirprobe{}) constructs a score matrix $M \in \mathbb{R}^{N \times N}$ for which each entry corresponds to a word pair's negative structural distance $-d_B(\boldsymbol{h}_i, \boldsymbol{h}_j)$. The shallowest node in the depth subspace $C$ is set as root. Entries in $M$ which correspond to an edge between $w_i$ and $w_j$ for which the word depths follow $\lVert C\boldsymbol{h}_i\rVert^2_2 > \lVert C\boldsymbol{h}_j\rVert^2_2$ are set to $-\infty$. A word's depth in subspace $C$ therefore corresponds to edge directionality. The directed graph is built from $M$ using Chu-Liu-Edmonds decoding \citep{chu1965,edmonds1967}.

\dirprobe{} extracts directed dependency parse trees, however it would require additional complexity to label each edge with a relation (e.g.\ using an additional probe). In the following, we propose a probe which can extract both directionality and relations while using fewer parameters and no dynamic programming-based graph-decoding algorithm.

\subsection{Relational Probing}\label{sec:relational-probing}

The incoming edge of each word $w_i$ is governed by a single relation. As such the task of dependency relation classification with $l$ relations can be simplified to a labeling task using a linear transformation $L \in \mathbb{R}^{e \times l}$ for which the probability of a word's relation $r_i$ being of class $l_k$ is given by:

\begin{equation}\label{eq:relation-classifier}
    \pdftooltip{p(r_i=l_k|w_i) = \text{softmax}(L\boldsymbol{h}_i)_k}{Screenreader Caption: Equation: p of r subscript i equals l subscript k given w subscript i equals k-th index of softmax of L times h subscript i.}
\end{equation}

and optimization uses standard cross-entropy loss given the gold label $r^\ast_i$ for each word $w_i$:

\begin{equation}\label{eq:xe-loss}
    \pdftooltip{\mathcal{L}_L(s) = - \frac{1}{N} \sum_{i=0}^N \ln{p(r^\ast_i|w_i)}\hspace{.3em}\text{.}}{L subscript L of s equals minus 1 over N times sum from index i equals 0 to N over the natural logarithm of p of r star subscript i given w subscript i.}
\end{equation}

Should dependency relations be encoded in contextualized embeddings, each dimension of the subspace $L$ will correspond to the prevalence of information relevant to each relation, quantifiable using relation classification accuracy (RelAcc).

\subsection{Constructing Dependency Parse Trees}\label{sec:depprobe-method}

\begin{algorithm}[t]
\SetAlgoLined
\textbf{input} Distance matrix $D_B \in \mathbb{R}^{N \times N}$, $p(l_k|w_i)$ of relation label $l_k$ given $w_i$\\
$w_r \leftarrow \argmax\limits_{w_i} p(\text{root}|w_i)$\\
$\mathcal{T}_w \leftarrow \{w_r\}, \mathcal{T}_e \leftarrow \{\}$\\
\While{$|\mathcal{T}_w| < N$}{
    $w_i, w_j \leftarrow \argmin\limits_{w_i, w_j} D_B(w_i \in \mathcal{T}_w, w_j)$\\
    $r_j \leftarrow \argmax\limits_{l_k} p(l_k|w_j) \text{ with } l_k \neq \text{root}$\\
    $\mathcal{T}_w \leftarrow \mathcal{T}_w \cup \{w_j\}$\\ $\mathcal{T}_e \leftarrow \mathcal{T}_e \cup \{(\overrightarrow{w_i, w_j}, r_j)\}$
}
\textbf{return} $\mathcal{T}_e$
 \caption{\depprobe{} Inference}
 \label{alg:depprobe}
\end{algorithm}

Combining structural probing (Section \ref{sec:undirected-probing}) and dependency relation probing (Section \ref{sec:relational-probing}), we propose a new probe for extracting fully directed and labeled dependency trees (\depprobe{}). It combines undirected graphs and relational information in a computationally efficient manner, adding labels while requiring \textit{less} parameters than prior unlabeled or multi-layer-perceptron-based approaches.

As outlined in Algorithm \ref{alg:depprobe} and illustrated in Figure~\ref{fig:depprobe}, \depprobe{} uses the distance matrix $D_B$ derived from the structural probe $B$ in conjunction with the relation probabilities of the relational probe $L$ (line 1). The graph is first rooted using the word $w_r$ for which $p(\text{root}|w_r)$ is highest (line 2). Iterating over the remaining words until all $w_j$ are covered in $\mathcal{T}_w$, an edge is drawn to each word $w_j$ from its head $w_i$ based on the minimum distance in $D_B$. The relation $r_j$ for an edge $(\overrightarrow{w_i, w_j}, r_j)$ is determined by taking the relation label $l_k$ which maximizes $p(r_j = l_k|w_j)$ with $l_k \neq \text{root}$ (line 6). The edge is then added to the set of labeled tree edges $\mathcal{T}_e$. With edge directionality being inferred as simply pointing away from the root, this procedure produces a dependency graph that is both directed and labeled without the need for additional complexity, running in $\mathcal{O}(n^2)$ while dynamic programming-based decoding such as \dirprobe{} have runtimes of up to $\mathcal{O}(n^3)$ \citep{stanojevic-cohen-2021-root}.

Constructing dependency trees from untuned embeddings requires the matrices $B$ and $L$, totaling $e \cdot b + e \cdot l$ trainable parameters. Optimization can be performed using gradient descent on the sum of losses $\mathcal{L}_B + \mathcal{L}_L$. With $l =$ 37 relations in UD, this constitutes a substantially reduced training effort compared to prior probing approaches (with subspace dimensionalities $b$ and $c$ typically set to 128) and multiple magnitudes fewer fine-tuned parameters than for a full biaffine attention parser.

\section{Experiments}\label{sec:experiments}

\subsection{Setup}

\paragraph{Parsers} In our experiments, we use the deep biaffine attention parser (\bap{}) by \citet{dozat2017} as implemented in \citet{van-der-goot-etal-2021-massive} as an upper bound for MLM-based parsing performance. As it is closest to our work, we further reimplement \dirprobe{} \citep{kulmizev-etal-2020-neural} with $b =$ 128 and $c =$ 128. Note that this approach produces directed, but unlabeled dependency graphs. Finally, we compare both methods to our directed and labeled probing approach, \depprobe{} with $b =$ 128 and $l =$ 37.

All methods use mBERT \citep{devlin-etal-2019-bert} as their encoder ($e =$ 768). For \bap{}, training the model includes fine-tuning the encoder's parameters, while for both probes they remain fixed and only the linear transformations are adjusted. This results in 183M tuned parameters for \bap{}, 197k for \dirprobe{} and 127k for \depprobe{}. Hyperparameters are set to the values reported by the authors,\footnote{For better comparability, we use the best single layer reported by \citet{kulmizev-etal-2020-neural} instead of the weighted sum over all layers.} while for \depprobe{} we perform an initial tuning step in Section \ref{sec:depprobe-tuning}.

\paragraph{Target Treebanks} As targets, we use the set of 13 treebanks proposed by \citet{kulmizev-etal-2019-deep}, using versions from Universal Dependencies v2.8 \citep{ud28}. They are diverse with respect to language family, morphological complexity and script (Appendix \ref{sec:experiment-setup}). This set further includes EN-EWT \citep{silveira-etal-2014-gold} which has been used in prior probing work for hyperparameter tuning, allowing us to tune \depprobe{} on the same data.

\paragraph{Metrics} We report labeled attachment scores (LAS) wherever possible (\bap{}, \depprobe{}) and unlabeled attachment scores (UAS) for all methods. For \depprobe{}'s hyperparameters, we evaluate undirected, unlabeled attachment scores (UUAS) as well as relation classification accuracy (RelAcc). One notable difference to prior work is that we include punctuation both during training and evaluation --- contrary to prior probing work which excludes all punctuation \citep{hewitt2019,kulmizev-etal-2020-neural,maudslay2020} --- since we are interested in the full parsing task.

\paragraph{Training} Each method is trained on each target treebank's training split and is evaluated on the test split. For cross-lingual transfer, models trained on one language are evaluated on the test splits of all other languages without any further tuning. For \depprobe{} tuning (Section \ref{sec:depprobe-tuning}) we use the development split of EN-EWT.

\bap{} uses the training schedule implemented in \citet{van-der-goot-etal-2021-massive} while \dirprobe{} and \depprobe{} use AdamW \citep{loshchilov2018decoupled} with a learning rate of $10^{-3}$ which is reduced by a factor of 10 each time the loss plateaus (see also \citealp{hewitt2019}).

Both probing methods are implemented using PyTorch \citep{pytorch} and use mBERT as implemented in the Transformers library \citep{wolf-etal-2020-transformers}. Each model is trained with three random initializations of which we report the mean.

\subsection{\depprobe{} Tuning}\label{sec:depprobe-tuning}

\begin{figure}
    \centering
    \hspace{-1em}
    \pdftooltip{\includegraphics[width=.5\textwidth]{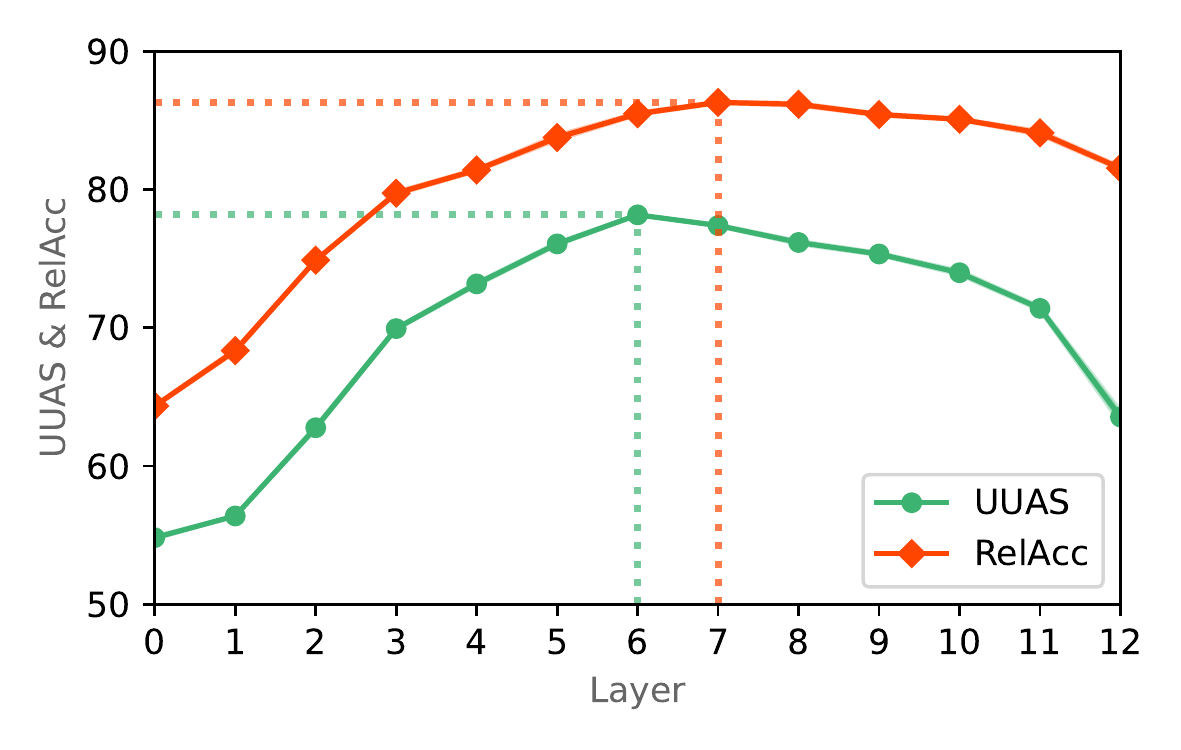}}{Screenreader Caption: UUAS: 54.8 at layer 0, 56.4 at layer 1, 62.8 at layer 2, 69.9 at layer 3, 73.2 at layer 4, 76.1 at layer 5, 78.2 at layer 6, 77.4 at layer 7, 76.2 at layer 8, 75.3 at layer 9, 74.0 at layer 10, 71.4 at layer 11, 63.5 at layer 12. RelAcc: 64.3 at layer 0, 68.3 at layer 1, 74.9 at layer 2, 79.7 at layer 3, 81.4 at layer 4, 83.8 at layer 5, 85.5 at layer 6, 86.3 at layer 7, 86.2 at layer 8, 85.4 at layer 9, 85.1 at layer 10, 84.1 at layer 11, 81.5 at layer 12.}
    \caption{\textbf{Layer-wise Performance on EWT (Dev)} for \depprobe{} as measured by UUAS for the structural probe $B$ and RelAcc for the relational probe $L$.}
    \label{fig:depprobe-layers}
\end{figure}

\begin{figure*}
    \centering
    \begin{subfigure}[b]{.3\textwidth}
        \pdftooltip{\includegraphics[width=\textwidth]{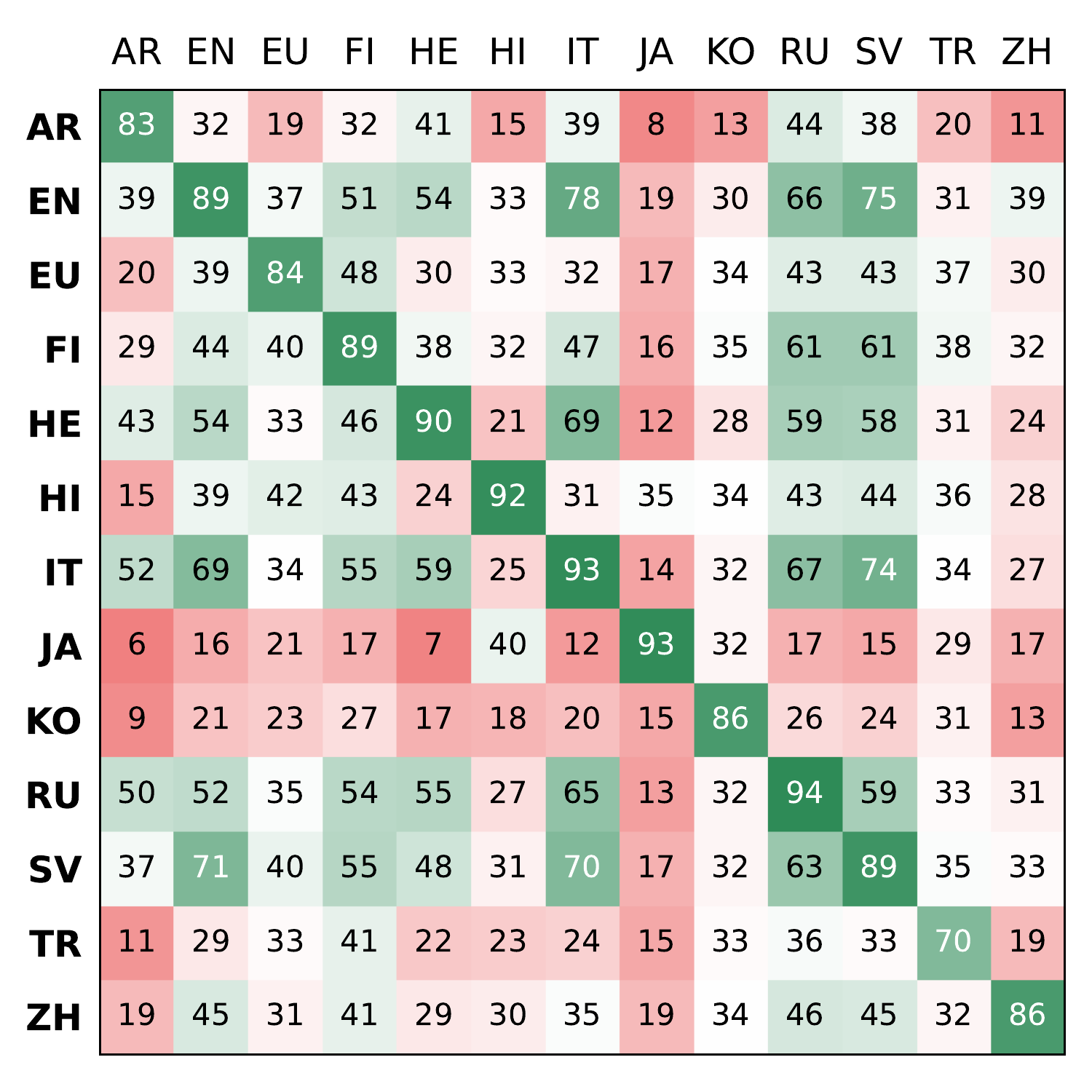}}{Screenreader Caption: AR to AR: 83, AR to EN: 32, AR to EU: 19, AR to FI: 32, AR to HE: 41, AR to HI: 15, AR to IT: 39, AR to JA: 8, AR to KO: 13, AR to RU: 44, AR to SV: 38, AR to TR: 20, AR to ZH: 11, EN to AR: 39, EN to EN: 89, EN to EU: 37, EN to FI: 51, EN to HE: 54, EN to HI: 33, EN to IT: 78, EN to JA: 19, EN to KO: 30, EN to RU: 66, EN to SV: 75, EN to TR: 31, EN to ZH: 39, EU to AR: 20, EU to EN: 39, EU to EU: 84, EU to FI: 48, EU to HE: 30, EU to HI: 33, EU to IT: 32, EU to JA: 17, EU to KO: 34, EU to RU: 43, EU to SV: 43, EU to TR: 37, EU to ZH: 30, FI to AR: 29, FI to EN: 44, FI to EU: 40, FI to FI: 89, FI to HE: 38, FI to HI: 32, FI to IT: 47, FI to JA: 16, FI to KO: 35, FI to RU: 61, FI to SV: 61, FI to TR: 38, FI to ZH: 32, HE to AR: 43, HE to EN: 54, HE to EU: 33, HE to FI: 46, HE to HE: 90, HE to HI: 21, HE to IT: 69, HE to JA: 12, HE to KO: 28, HE to RU: 59, HE to SV: 58, HE to TR: 31, HE to ZH: 24, HI to AR: 15, HI to EN: 39, HI to EU: 42, HI to FI: 43, HI to HE: 24, HI to HI: 92, HI to IT: 31, HI to JA: 35, HI to KO: 34, HI to RU: 43, HI to SV: 44, HI to TR: 36, HI to ZH: 28, IT to AR: 52, IT to EN: 69, IT to EU: 34, IT to FI: 55, IT to HE: 59, IT to HI: 25, IT to IT: 93, IT to JA: 14, IT to KO: 32, IT to RU: 67, IT to SV: 74, IT to TR: 34, IT to ZH: 27, JA to AR: 6, JA to EN: 16, JA to EU: 21, JA to FI: 17, JA to HE: 7, JA to HI: 40, JA to IT: 12, JA to JA: 93, JA to KO: 32, JA to RU: 17, JA to SV: 15, JA to TR: 29, JA to ZH: 17, KO to AR: 9, KO to EN: 21, KO to EU: 23, KO to FI: 27, KO to HE: 17, KO to HI: 18, KO to IT: 20, KO to JA: 15, KO to KO: 86, KO to RU: 26, KO to SV: 24, KO to TR: 31, KO to ZH: 13, RU to AR: 50, RU to EN: 52, RU to EU: 35, RU to FI: 54, RU to HE: 55, RU to HI: 27, RU to IT: 65, RU to JA: 13, RU to KO: 32, RU to RU: 94, RU to SV: 59, RU to TR: 33, RU to ZH: 31, SV to AR: 37, SV to EN: 71, SV to EU: 40, SV to FI: 55, SV to HE: 48, SV to HI: 31, SV to IT: 70, SV to JA: 17, SV to KO: 32, SV to RU: 63, SV to SV: 89, SV to TR: 35, SV to ZH: 33, TR to AR: 11, TR to EN: 29, TR to EU: 33, TR to FI: 41, TR to HE: 22, TR to HI: 23, TR to IT: 24, TR to JA: 15, TR to KO: 33, TR to RU: 36, TR to SV: 33, TR to TR: 70, TR to ZH: 19, ZH to AR: 19, ZH to EN: 45, ZH to EU: 31, ZH to FI: 41, ZH to HE: 29, ZH to HI: 30, ZH to IT: 35, ZH to JA: 19, ZH to KO: 34, ZH to RU: 46, ZH to SV: 45, ZH to TR: 32, ZH to ZH: 86.}
        \caption{\bap{} (LAS)}
        \vspace{.9em}
        \label{fig:las-bap}
    \end{subfigure}
    \begin{subfigure}[b]{.3\textwidth}
        \pdftooltip{\includegraphics[width=\textwidth]{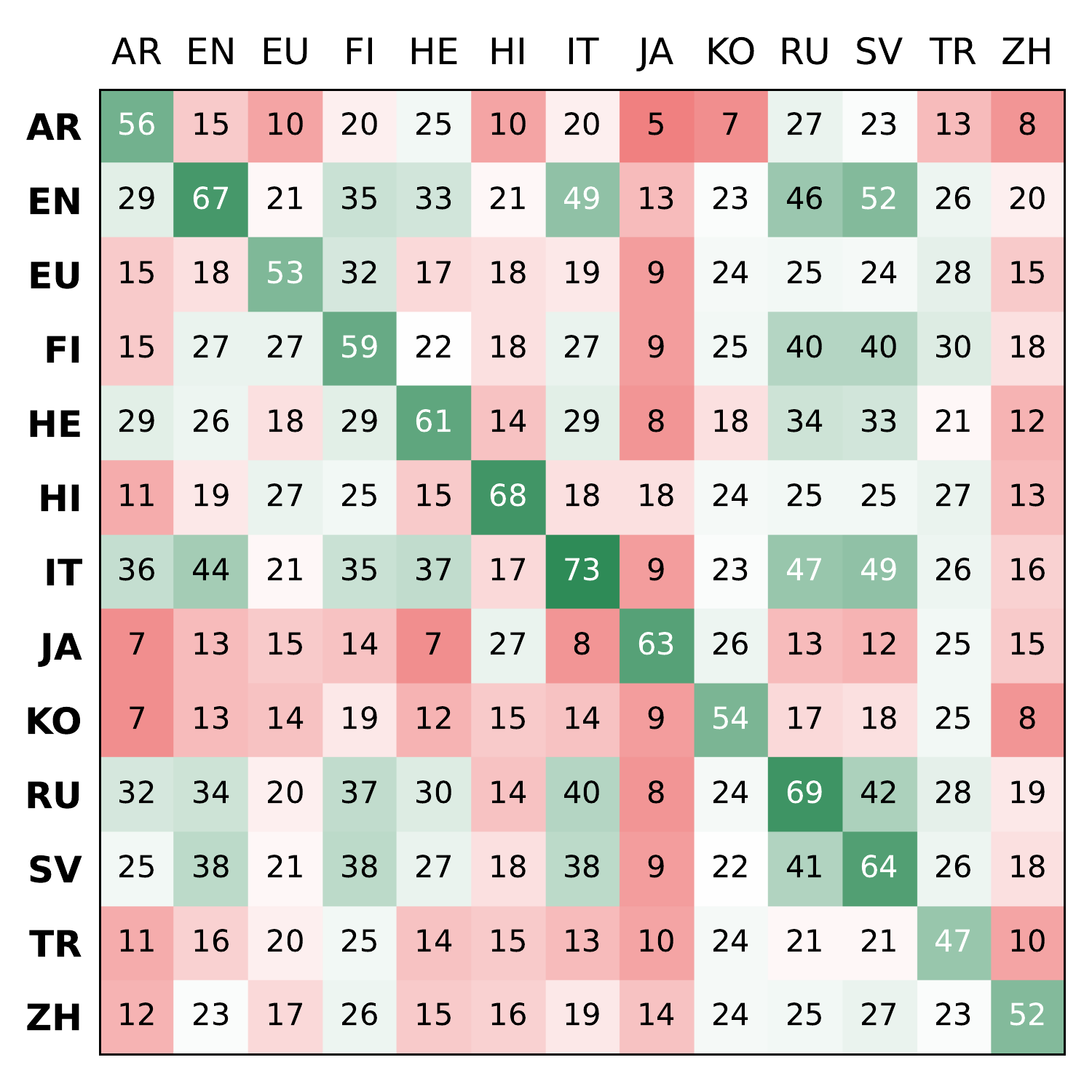}}{Screenreader Caption: AR to AR: 56, AR to EN: 15, AR to EU: 10, AR to FI: 20, AR to HE: 25, AR to HI: 10, AR to IT: 20, AR to JA: 5, AR to KO: 7, AR to RU: 27, AR to SV: 23, AR to TR: 13, AR to ZH: 8, EN to AR: 29, EN to EN: 67, EN to EU: 21, EN to FI: 35, EN to HE: 33, EN to HI: 21, EN to IT: 49, EN to JA: 13, EN to KO: 23, EN to RU: 46, EN to SV: 52, EN to TR: 26, EN to ZH: 20, EU to AR: 15, EU to EN: 18, EU to EU: 53, EU to FI: 32, EU to HE: 17, EU to HI: 18, EU to IT: 19, EU to JA: 9, EU to KO: 24, EU to RU: 25, EU to SV: 24, EU to TR: 28, EU to ZH: 15, FI to AR: 15, FI to EN: 27, FI to EU: 27, FI to FI: 59, FI to HE: 22, FI to HI: 18, FI to IT: 27, FI to JA: 9, FI to KO: 25, FI to RU: 40, FI to SV: 40, FI to TR: 30, FI to ZH: 18, HE to AR: 29, HE to EN: 26, HE to EU: 18, HE to FI: 29, HE to HE: 61, HE to HI: 14, HE to IT: 29, HE to JA: 8, HE to KO: 18, HE to RU: 34, HE to SV: 33, HE to TR: 21, HE to ZH: 12, HI to AR: 11, HI to EN: 19, HI to EU: 27, HI to FI: 25, HI to HE: 15, HI to HI: 68, HI to IT: 18, HI to JA: 18, HI to KO: 24, HI to RU: 25, HI to SV: 25, HI to TR: 27, HI to ZH: 13, IT to AR: 36, IT to EN: 44, IT to EU: 21, IT to FI: 35, IT to HE: 37, IT to HI: 17, IT to IT: 73, IT to JA: 9, IT to KO: 23, IT to RU: 47, IT to SV: 49, IT to TR: 26, IT to ZH: 16, JA to AR: 7, JA to EN: 13, JA to EU: 15, JA to FI: 14, JA to HE: 7, JA to HI: 27, JA to IT: 8, JA to JA: 63, JA to KO: 26, JA to RU: 13, JA to SV: 12, JA to TR: 25, JA to ZH: 15, KO to AR: 7, KO to EN: 13, KO to EU: 14, KO to FI: 19, KO to HE: 12, KO to HI: 15, KO to IT: 14, KO to JA: 9, KO to KO: 54, KO to RU: 17, KO to SV: 18, KO to TR: 25, KO to ZH: 8, RU to AR: 32, RU to EN: 34, RU to EU: 20, RU to FI: 37, RU to HE: 30, RU to HI: 14, RU to IT: 40, RU to JA: 8, RU to KO: 24, RU to RU: 69, RU to SV: 42, RU to TR: 28, RU to ZH: 19, SV to AR: 25, SV to EN: 38, SV to EU: 21, SV to FI: 38, SV to HE: 27, SV to HI: 18, SV to IT: 38, SV to JA: 9, SV to KO: 22, SV to RU: 41, SV to SV: 64, SV to TR: 26, SV to ZH: 18, TR to AR: 11, TR to EN: 16, TR to EU: 20, TR to FI: 25, TR to HE: 14, TR to HI: 15, TR to IT: 13, TR to JA: 10, TR to KO: 24, TR to RU: 21, TR to SV: 21, TR to TR: 47, TR to ZH: 10, ZH to AR: 12, ZH to EN: 23, ZH to EU: 17, ZH to FI: 26, ZH to HE: 15, ZH to HI: 16, ZH to IT: 19, ZH to JA: 14, ZH to KO: 24, ZH to RU: 25, ZH to SV: 27, ZH to TR: 23, ZH to ZH: 52.}
        \caption{\depprobe{} (LAS)}
        \vspace{.9em}
        \label{fig:las-dirprobe}
    \end{subfigure}
    \begin{subfigure}[b]{.3\textwidth}
        \begin{center}
        \resizebox{.85\textwidth}{!}{
        \begin{tabular}{lrrr}
        \toprule
        \textsc{Model} & \bap{} & \textsc{Dep} & \textsc{Dir} \\
        \midrule
        \multirow{2}*{\textsc{LAS=l}\vspace{.4em}} & 88 & 60 & --- \\[-.3em]
        & {\footnotesize$\pm$6.4} & {\footnotesize$\pm$7.8} & \\
        \multirow{2}*{\textsc{LAS$\neg$l}\vspace{.4em}} & 35 & 22 & --- \\[-.3em]
        & {\footnotesize$\pm$15.7} & {\footnotesize$\pm$9.9} & \\
        \midrule
        \multirow{2}*{\textsc{UAS=l}\vspace{.4em}} & 91 & 67 & 70 \\[-.3em]
        & {\footnotesize$\pm$5.0} & {\footnotesize$\pm$6.7} & {\footnotesize$\pm$7.8} \\
        \multirow{2}*{\textsc{UAS$\neg$l}\vspace{.4em}} & 52 & 38 & 36 \\[-.3em]
        & {\footnotesize$\pm$14.5} & {\footnotesize$\pm$8.8} & {\footnotesize$\pm$10.4} \\
        \bottomrule
        \end{tabular}
        }
        \end{center}
        \vspace{1em}
        \caption{Mean in-language (\textsc{=l}) and transfer (\textsc{$\neg$l}) UAS/LAS ($\pm$ stddev).}
        \label{tab:transfer-results}
    \end{subfigure}
    \begin{subfigure}[m]{.3\textwidth}
        \pdftooltip{\includegraphics[width=\textwidth]{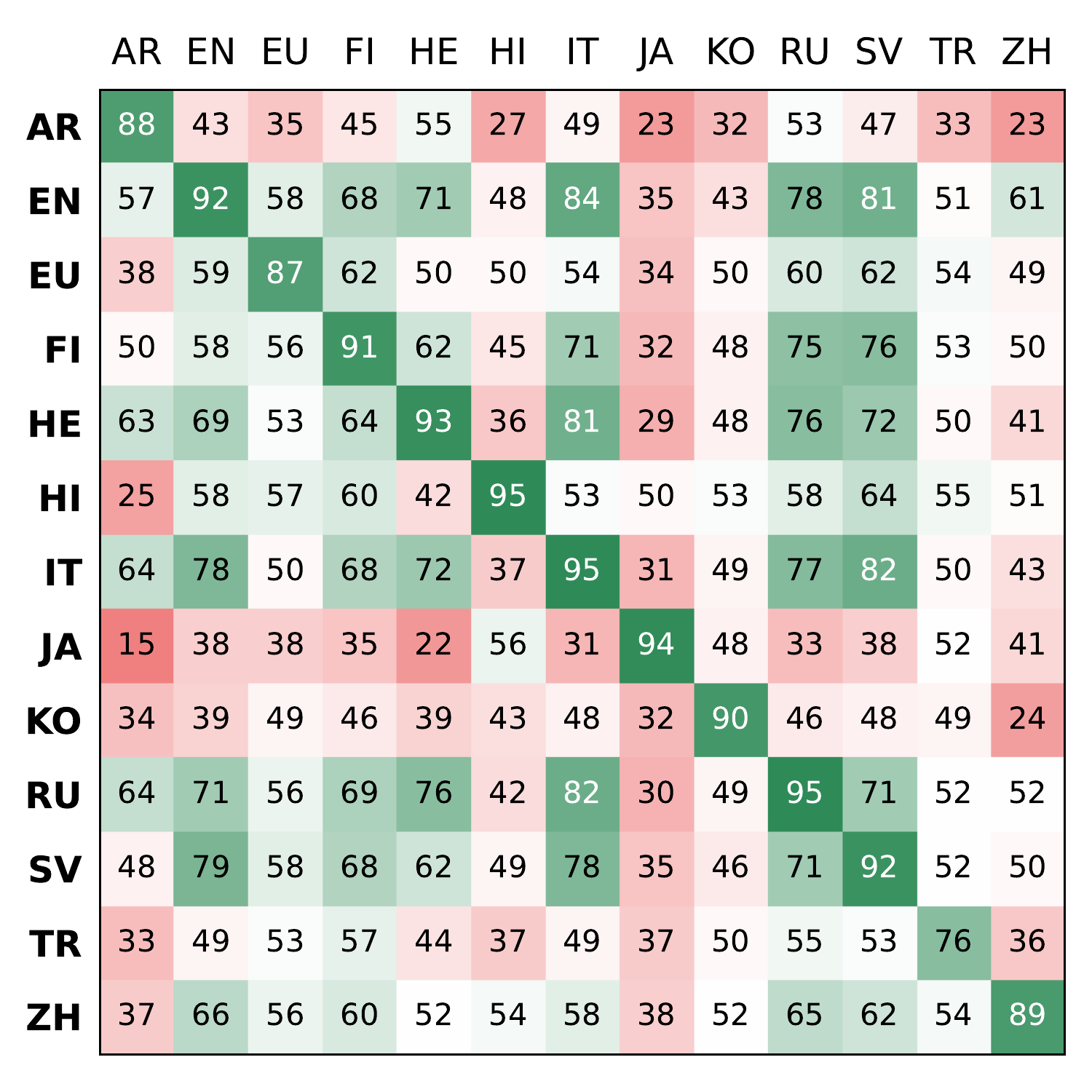}}{Screenreader Caption: AR to AR: 88, AR to EN: 43, AR to EU: 35, AR to FI: 45, AR to HE: 55, AR to HI: 27, AR to IT: 49, AR to JA: 23, AR to KO: 32, AR to RU: 53, AR to SV: 47, AR to TR: 33, AR to ZH: 23, EN to AR: 57, EN to EN: 92, EN to EU: 58, EN to FI: 68, EN to HE: 71, EN to HI: 48, EN to IT: 84, EN to JA: 35, EN to KO: 43, EN to RU: 78, EN to SV: 81, EN to TR: 51, EN to ZH: 61, EU to AR: 38, EU to EN: 59, EU to EU: 87, EU to FI: 62, EU to HE: 50, EU to HI: 50, EU to IT: 54, EU to JA: 34, EU to KO: 50, EU to RU: 60, EU to SV: 62, EU to TR: 54, EU to ZH: 49, FI to AR: 50, FI to EN: 58, FI to EU: 56, FI to FI: 91, FI to HE: 62, FI to HI: 45, FI to IT: 71, FI to JA: 32, FI to KO: 48, FI to RU: 75, FI to SV: 76, FI to TR: 53, FI to ZH: 50, HE to AR: 63, HE to EN: 69, HE to EU: 53, HE to FI: 64, HE to HE: 93, HE to HI: 36, HE to IT: 81, HE to JA: 29, HE to KO: 48, HE to RU: 76, HE to SV: 72, HE to TR: 50, HE to ZH: 41, HI to AR: 25, HI to EN: 58, HI to EU: 57, HI to FI: 60, HI to HE: 42, HI to HI: 95, HI to IT: 53, HI to JA: 50, HI to KO: 53, HI to RU: 58, HI to SV: 64, HI to TR: 55, HI to ZH: 51, IT to AR: 64, IT to EN: 78, IT to EU: 50, IT to FI: 68, IT to HE: 72, IT to HI: 37, IT to IT: 95, IT to JA: 31, IT to KO: 49, IT to RU: 77, IT to SV: 82, IT to TR: 50, IT to ZH: 43, JA to AR: 15, JA to EN: 38, JA to EU: 38, JA to FI: 35, JA to HE: 22, JA to HI: 56, JA to IT: 31, JA to JA: 94, JA to KO: 48, JA to RU: 33, JA to SV: 38, JA to TR: 52, JA to ZH: 41, KO to AR: 34, KO to EN: 39, KO to EU: 49, KO to FI: 46, KO to HE: 39, KO to HI: 43, KO to IT: 48, KO to JA: 32, KO to KO: 90, KO to RU: 46, KO to SV: 48, KO to TR: 49, KO to ZH: 24, RU to AR: 64, RU to EN: 71, RU to EU: 56, RU to FI: 69, RU to HE: 76, RU to HI: 42, RU to IT: 82, RU to JA: 30, RU to KO: 49, RU to RU: 95, RU to SV: 71, RU to TR: 52, RU to ZH: 52, SV to AR: 48, SV to EN: 79, SV to EU: 58, SV to FI: 68, SV to HE: 62, SV to HI: 49, SV to IT: 78, SV to JA: 35, SV to KO: 46, SV to RU: 71, SV to SV: 92, SV to TR: 52, SV to ZH: 50, TR to AR: 33, TR to EN: 49, TR to EU: 53, TR to FI: 57, TR to HE: 44, TR to HI: 37, TR to IT: 49, TR to JA: 37, TR to KO: 50, TR to RU: 55, TR to SV: 53, TR to TR: 76, TR to ZH: 36, ZH to AR: 37, ZH to EN: 66, ZH to EU: 56, ZH to FI: 60, ZH to HE: 52, ZH to HI: 54, ZH to IT: 58, ZH to JA: 38, ZH to KO: 52, ZH to RU: 65, ZH to SV: 62, ZH to TR: 54, ZH to ZH: 89.}
        \caption{\bap{} (UAS)}
        \label{fig:uas-bap}
    \end{subfigure}
    \begin{subfigure}[m]{.3\textwidth}
        \pdftooltip{\includegraphics[width=\textwidth]{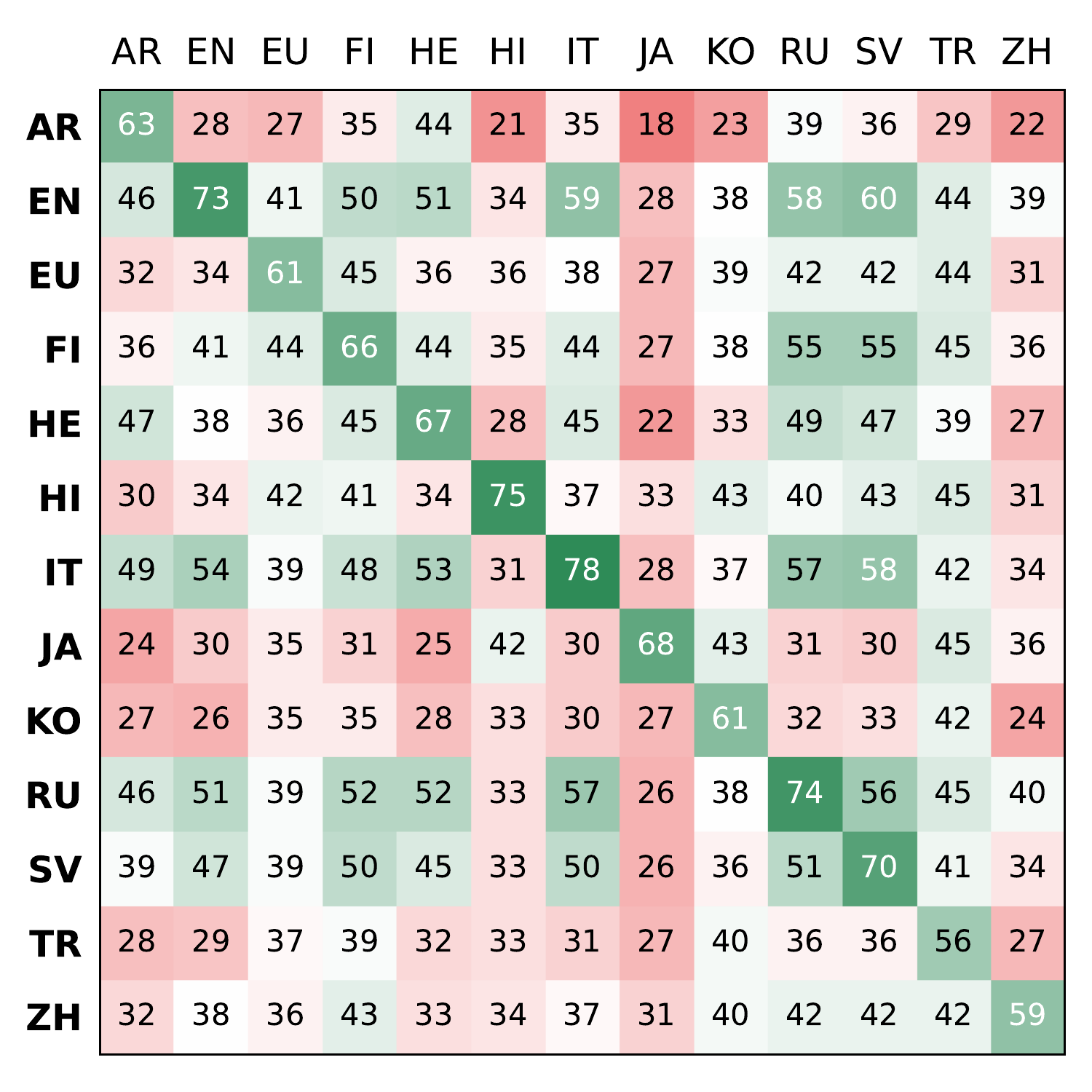}}{Screenreader Caption: AR to AR: 63, AR to EN: 28, AR to EU: 27, AR to FI: 35, AR to HE: 44, AR to HI: 21, AR to IT: 35, AR to JA: 18, AR to KO: 23, AR to RU: 39, AR to SV: 36, AR to TR: 29, AR to ZH: 22, EN to AR: 46, EN to EN: 73, EN to EU: 41, EN to FI: 50, EN to HE: 51, EN to HI: 34, EN to IT: 59, EN to JA: 28, EN to KO: 38, EN to RU: 58, EN to SV: 60, EN to TR: 44, EN to ZH: 39, EU to AR: 32, EU to EN: 34, EU to EU: 61, EU to FI: 45, EU to HE: 36, EU to HI: 36, EU to IT: 38, EU to JA: 27, EU to KO: 39, EU to RU: 42, EU to SV: 42, EU to TR: 44, EU to ZH: 31, FI to AR: 36, FI to EN: 41, FI to EU: 44, FI to FI: 66, FI to HE: 44, FI to HI: 35, FI to IT: 44, FI to JA: 27, FI to KO: 38, FI to RU: 55, FI to SV: 55, FI to TR: 45, FI to ZH: 36, HE to AR: 47, HE to EN: 38, HE to EU: 36, HE to FI: 45, HE to HE: 67, HE to HI: 28, HE to IT: 45, HE to JA: 22, HE to KO: 33, HE to RU: 49, HE to SV: 47, HE to TR: 39, HE to ZH: 27, HI to AR: 30, HI to EN: 34, HI to EU: 42, HI to FI: 41, HI to HE: 34, HI to HI: 75, HI to IT: 37, HI to JA: 33, HI to KO: 43, HI to RU: 40, HI to SV: 43, HI to TR: 45, HI to ZH: 31, IT to AR: 49, IT to EN: 54, IT to EU: 39, IT to FI: 48, IT to HE: 53, IT to HI: 31, IT to IT: 78, IT to JA: 28, IT to KO: 37, IT to RU: 57, IT to SV: 58, IT to TR: 42, IT to ZH: 34, JA to AR: 24, JA to EN: 30, JA to EU: 35, JA to FI: 31, JA to HE: 25, JA to HI: 42, JA to IT: 30, JA to JA: 68, JA to KO: 43, JA to RU: 31, JA to SV: 30, JA to TR: 45, JA to ZH: 36, KO to AR: 27, KO to EN: 26, KO to EU: 35, KO to FI: 35, KO to HE: 28, KO to HI: 33, KO to IT: 30, KO to JA: 27, KO to KO: 61, KO to RU: 32, KO to SV: 33, KO to TR: 42, KO to ZH: 24, RU to AR: 46, RU to EN: 51, RU to EU: 39, RU to FI: 52, RU to HE: 52, RU to HI: 33, RU to IT: 57, RU to JA: 26, RU to KO: 38, RU to RU: 74, RU to SV: 56, RU to TR: 45, RU to ZH: 40, SV to AR: 39, SV to EN: 47, SV to EU: 39, SV to FI: 50, SV to HE: 45, SV to HI: 33, SV to IT: 50, SV to JA: 26, SV to KO: 36, SV to RU: 51, SV to SV: 70, SV to TR: 41, SV to ZH: 34, TR to AR: 28, TR to EN: 29, TR to EU: 37, TR to FI: 39, TR to HE: 32, TR to HI: 33, TR to IT: 31, TR to JA: 27, TR to KO: 40, TR to RU: 36, TR to SV: 36, TR to TR: 56, TR to ZH: 27, ZH to AR: 32, ZH to EN: 38, ZH to EU: 36, ZH to FI: 43, ZH to HE: 33, ZH to HI: 34, ZH to IT: 37, ZH to JA: 31, ZH to KO: 40, ZH to RU: 42, ZH to SV: 42, ZH to TR: 42, ZH to ZH: 59.}
        \caption{\depprobe{} (UAS)}
        \label{fig:uas-depprobe}
    \end{subfigure}
    \begin{subfigure}[m]{.3\textwidth}
        \pdftooltip{\includegraphics[width=\textwidth]{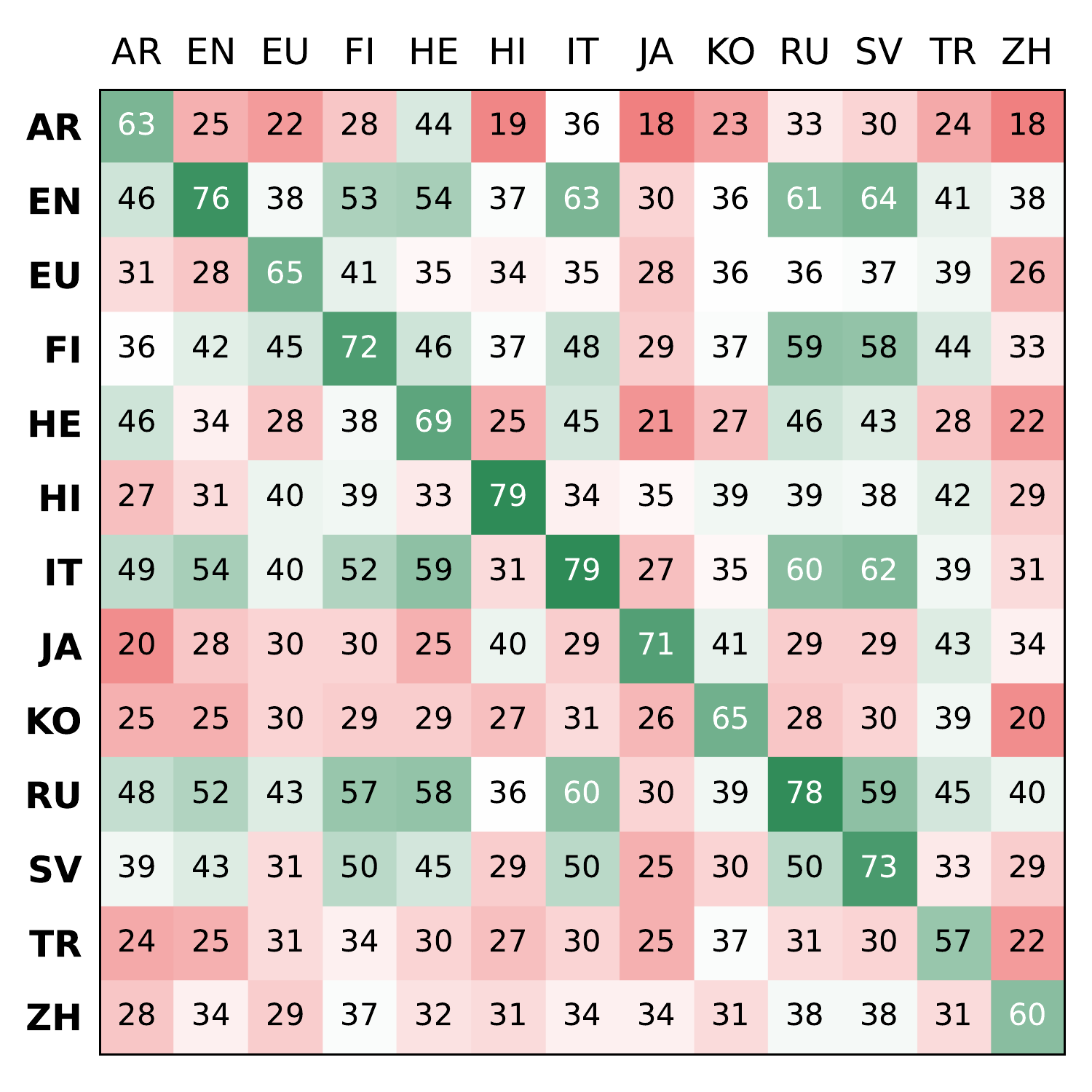}}{Screenreader Caption: AR to AR: 63, AR to EN: 25, AR to EU: 22, AR to FI: 28, AR to HE: 44, AR to HI: 19, AR to IT: 36, AR to JA: 18, AR to KO: 23, AR to RU: 33, AR to SV: 30, AR to TR: 24, AR to ZH: 18, EN to AR: 46, EN to EN: 76, EN to EU: 38, EN to FI: 53, EN to HE: 54, EN to HI: 37, EN to IT: 63, EN to JA: 30, EN to KO: 36, EN to RU: 61, EN to SV: 64, EN to TR: 41, EN to ZH: 38, EU to AR: 31, EU to EN: 28, EU to EU: 65, EU to FI: 41, EU to HE: 35, EU to HI: 34, EU to IT: 35, EU to JA: 28, EU to KO: 36, EU to RU: 36, EU to SV: 37, EU to TR: 39, EU to ZH: 26, FI to AR: 36, FI to EN: 42, FI to EU: 45, FI to FI: 72, FI to HE: 46, FI to HI: 37, FI to IT: 48, FI to JA: 29, FI to KO: 37, FI to RU: 59, FI to SV: 58, FI to TR: 44, FI to ZH: 33, HE to AR: 46, HE to EN: 34, HE to EU: 28, HE to FI: 38, HE to HE: 69, HE to HI: 25, HE to IT: 45, HE to JA: 21, HE to KO: 27, HE to RU: 46, HE to SV: 43, HE to TR: 28, HE to ZH: 22, HI to AR: 27, HI to EN: 31, HI to EU: 40, HI to FI: 39, HI to HE: 33, HI to HI: 79, HI to IT: 34, HI to JA: 35, HI to KO: 39, HI to RU: 39, HI to SV: 38, HI to TR: 42, HI to ZH: 29, IT to AR: 49, IT to EN: 54, IT to EU: 40, IT to FI: 52, IT to HE: 59, IT to HI: 31, IT to IT: 79, IT to JA: 27, IT to KO: 35, IT to RU: 60, IT to SV: 62, IT to TR: 39, IT to ZH: 31, JA to AR: 20, JA to EN: 28, JA to EU: 30, JA to FI: 30, JA to HE: 25, JA to HI: 40, JA to IT: 29, JA to JA: 71, JA to KO: 41, JA to RU: 29, JA to SV: 29, JA to TR: 43, JA to ZH: 34, KO to AR: 25, KO to EN: 25, KO to EU: 30, KO to FI: 29, KO to HE: 29, KO to HI: 27, KO to IT: 31, KO to JA: 26, KO to KO: 65, KO to RU: 28, KO to SV: 30, KO to TR: 39, KO to ZH: 20, RU to AR: 48, RU to EN: 52, RU to EU: 43, RU to FI: 57, RU to HE: 58, RU to HI: 36, RU to IT: 60, RU to JA: 30, RU to KO: 39, RU to RU: 78, RU to SV: 59, RU to TR: 45, RU to ZH: 40, SV to AR: 39, SV to EN: 43, SV to EU: 31, SV to FI: 50, SV to HE: 45, SV to HI: 29, SV to IT: 50, SV to JA: 25, SV to KO: 30, SV to RU: 50, SV to SV: 73, SV to TR: 33, SV to ZH: 29, TR to AR: 24, TR to EN: 25, TR to EU: 31, TR to FI: 34, TR to HE: 30, TR to HI: 27, TR to IT: 30, TR to JA: 25, TR to KO: 37, TR to RU: 31, TR to SV: 30, TR to TR: 57, TR to ZH: 22, ZH to AR: 28, ZH to EN: 34, ZH to EU: 29, ZH to FI: 37, ZH to HE: 32, ZH to HI: 31, ZH to IT: 34, ZH to JA: 34, ZH to KO: 31, ZH to RU: 38, ZH to SV: 38, ZH to TR: 31, ZH to ZH: 60.}
        \caption{\dirprobe{} (UAS)}
        \label{fig:uas-dirprobe}
    \end{subfigure}
    \caption{\textbf{In-language and Cross-lingual Transfer Performance} for 13 target treebanks (\textbf{train} $\rightarrow$ test) in UAS for \bap{} (fully tuned parser), \depprobe{}, \dirprobe{} and LAS for \bap{}, \depprobe{} (\dirprobe{} is unlabeled).}
    \label{fig:transfer-performance}
\end{figure*}

As prior work has repeatedly found that MLM layers encode different linguistic information, the layers which are most relevant for a probe's task are typically first identified \citep{tenney2019,hewitt2019}. Following this paradigm, we train \depprobe{} on embeddings from each layer of mBERT. Layer 0 is equivalent to the first, non-contextualized embeddings while layer 12 is the output of the last attention heads. The probe is trained on EN-EWT and evaluated on its development split using UUAS for the structural transformation $B$ (akin to \citealp{hewitt2019}) as well as RelAcc for the relational transformation $L$.

Figure \ref{fig:depprobe-layers} shows that structure is most prevalent around layer 6 at 78 UUAS, corroborating the 6--8 range identified by prior work \citep{tenney2019,hewitt2019,chi2020}. Dependency relations are easiest to retrieve at around layer 7 with an accuracy of 86\%. The standard deviation across initializations is around 0.1 in both cases. Based on these tuning results, we use layer 6 for structural probing and layer 7 for relational probing in the following experiments.

\subsection{Parsing Performance}

Figure \ref{fig:transfer-performance} lists UAS for all methods and LAS for \bap{} and \depprobe{} both on target-language test data (=L) and zero-shot transfer targets ($\neg$L). Table \ref{tab:transfer-results} further shows the mean results for each setting.

Unsurprisingly, the full parametrization of \bap{} performs best, with in-language scores of 88 LAS and 91 UAS. For zero-shot transfer, these scores drop to 35 LAS and 52 UAS, with some language pairs seeing differences of up to 85 points: e.g.\ JA $\rightarrow$ JA (93 LAS) versus AR $\rightarrow$ JA (8 LAS) in Figure \ref{fig:las-bap}. This again confirms the importance of selecting appropriate source data for any given target.

Both probes, with their limited parametrization, fall short of the full parser's performance, but still reach up to 73 LAS and 79 UAS. \dirprobe{} has a mean in-language UAS which is 3 points higher than for \depprobe{}, attributable to the more complex decoder. Due to \dirprobe{}'s output structures being unlabeled, we cannot compare LAS.

\depprobe{} reaches a competitive 67 UAS despite its much simpler decoding procedure and appears to be more stable for zero-shot transfer as it outperforms \dirprobe{} by around 2 UAS while maintaining a lower standard deviation. Most importantly, it produces directed and \textit{labeled} parses such that we can fully compare it to \bap{}. Considering that \depprobe{} has more than three orders of magnitude fewer tunable parameters, a mean in-language LAS of 60 is considerable and highlights the large degree of latent dependency information in untuned, contextual embeddings. For zero-shot transfer, the performance gap to \bap{} narrows to 13 LAS and 14 UAS.

\subsection{Transfer Prediction}\label{sec:transfer-prediction}

Given that \depprobe{} provides a highly parameter-efficient method for producing directed, labeled parse trees, we next investigate whether its performance patterns are indicative of the full parser's performance and could aid in selecting an appropriate source treebank for a given target without having to train the 183 million parameters of \bap{}.

\paragraph{Setup} Comparing UAS and LAS of \bap{} with respective scores of \depprobe{} and \dirprobe{}, we compute the Pearson correlation coefficient $\rho$ and the weighted Kendall's $\tau_w$ \citep{vigna2015}. The latter can be interpreted as corresponding to a correlation in $[-1,1]$, and that given a probe ranking one source treebank over another, the probability of this higher rank corresponding to higher performance in the full parser is $\frac{\tau_w + 1}{2}$. All reported correlations are significant at $p < 0.001$. Similarly, differences between correlation coefficients are also significant at $p < 0.001$ as measured using a standard Z-test.
In addition to the probes, we also compare against a method commonly employed by practitioners by using the cosine similarity of typological features from the URIEL database as represented in lang2vec (\citealp{littell-etal-2017-uriel}; \textsc{L2V}) between our 13 targets (details in Appendix \ref{sec:experiment-setup}).

\begin{table}
\centering
\resizebox{.4\textwidth}{!}{
\begin{tabular}{lrrrr}
\toprule
\multirow{2}*{\textsc{Model}} & \multicolumn{2}{c}{\textsc{LAS}} & \multicolumn{2}{c}{\textsc{UAS}} \\
& \multicolumn{1}{c}{$\rho$} & \multicolumn{1}{c}{$\tau_w$} & \multicolumn{1}{c}{$\rho$} & \multicolumn{1}{c}{$\tau_w$} \\
\midrule
\textsc{L2V} & .86 & .72 & .80 & .70 \\
\midrule
\dirprobe{} & --- & --- & .91 & .81 \\
\depprobe{} & \textbf{.97} & \textbf{.88} & .94 & .85 \\
\bottomrule
\end{tabular}
}
\caption{\label{tab:transfer-correlation} \textbf{Transfer Correlation with \bap{}.} Pearson $\rho$ and weighted Kendall's $\tau_w$ for \bap{}'s LAS and UAS with respect to \dirprobe{}'s UAS, \depprobe{}'s UAS and LAS as well as lang2vec cosine similarity (\textsc{L2V}).}
\end{table}

\paragraph{Results} Table \ref{tab:transfer-correlation} shows that the \textsc{L2V} baseline correlates with final parser performance, but that actual dependency parses yield significantly higher correlation and predictive power. For UAS, we find that despite having similar attachment scores, \depprobe{} performance correlates higher with \bap{} than that of \dirprobe{}, both with respect to predicting the ability to parse any particular language as well as ranking the best source to transfer from. Using the labeled parse trees of \depprobe{} results in almost perfect correlation with \bap{}'s LAS at $\rho =$ .97 as well as a $\tau_w$ of .88, highlighting the importance of modeling the full task and including dependency relation information. Using Kendall's $\tau_w$ with respect to LAS, we can estimate that selecting the highest performing source treebank from \depprobe{} to train the full parser will be the best choice 94\% of the time for any treebank pair.

\section{Analysis}\label{sec:analysis}

\subsection{Tree Depth versus Relations}\label{sec:ssa-analysis}

\begin{table}
\vspace{.35em}
\centering
\resizebox{.4\textwidth}{!}{
\begin{tabular}{lrrrr}
\toprule
\multirow{2}*{\textsc{Model}} & \multicolumn{2}{c}{\textsc{LAS}} & \multicolumn{2}{c}{\textsc{UAS}} \\
& \multicolumn{1}{c}{$\rho$} & \multicolumn{1}{c}{$\tau_w$} & \multicolumn{1}{c}{$\rho$} & \multicolumn{1}{c}{$\tau_w$} \\
\midrule
\textsc{SSA{\footnotesize -Struct}} & .68 & .42 & .60 & .43 \\
\textsc{SSA{\footnotesize -Depth}} & .62 & .34 & .53 & .35 \\
\textsc{SSA{\footnotesize -Rel}} & \textbf{.73} & \textbf{.55} & .65 & .53 \\
\bottomrule
\end{tabular}
}
\vspace{.35em}
\caption{\label{tab:ssa-correlation} \textbf{SSA Correlation with \bap{}.} Pearson $\rho$ and weighted Kendall's $\tau_w$ for \bap{}'s LAS and UAS with respect to subspace angles between structural (\textsc{Struct}), depth (\textsc{Depth}) and relation probes (\textsc{Rel}).}
\vspace{-.4em}
\end{table}

Why does \depprobe{} predict transfer performance more accurately than \dirprobe{} despite its simpler architecture? As each probe consists only of two matrices optimized to extract tree structural, depth or relational information, we can directly compare the similarity of all task-relevant parameters across languages against the full \bap{}'s cross-lingual performance.

In order to measure the similarity of probe matrices from different languages, we use mean subspace angles (\citealp{knyazev2002}; SSA), similarly to prior probing work \citep{chi2020}. Intuitively, SSA quantifies the energy required to transform one matrix to another by converting the singular values of the transformation into angles between 0\degree and 90\degree. SSAs are computed for the structural probe (\textsc{SSA{\footnotesize -Struct}}) which is equivalent in both methods, \dirprobe{}'s depth probe (\textsc{SSA{\footnotesize -Depth}}) and \depprobe{}'s relational probe (\textsc{SSA{\footnotesize -Rel}}). We use Pearson $\rho$ and the weighted Kendall's $\tau_w$ to measure the correlation between cross-lingual probe SSAs and \bap{} performance. This allows us to investigate which type of information is most important for final parsing performance.

From Table \ref{tab:ssa-correlation}, we can observe that SSAs between probes of different languages correlate less with transfer performance than UAS or LAS (Table \ref{tab:transfer-correlation}), underlining the importance of extracting full parses. Among the different types of dependency information, we observe that SSAs between the \textit{relational} probes used by \depprobe{} correlate highest with final performance at .73 for LAS and .65 for UAS. Structural probing correlates significantly both with \bap{}'s LAS and UAS at .68 and .60 respectively, but to a lesser degree. Probes for tree depth have the lowest correlation at .62 for LAS and .53 for UAS. Despite tree depth being a distinctive syntactic feature for language pairs such as the agglutinative Turkish and the more function word-based English, depth is either not as relevant for \bap{} or may be represented less consistently in embeddings across languages, leading to lower correlation between SSAs and final performance.

\subsection{Full Parser versus Probe}\label{sec:dependency-analysis}

\begin{figure*}
    \begin{subfigure}[m]{.96\textwidth}
        \pdftooltip{\includegraphics[width=\textwidth]{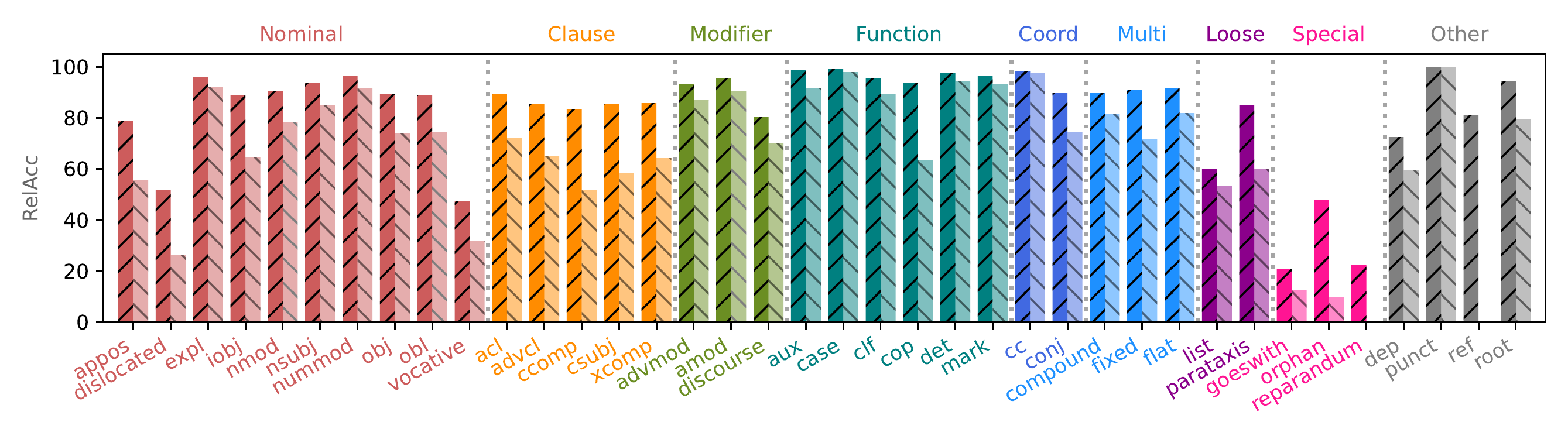}}{Screenreader Caption: nominal: appos: 78\% BAP \& 55\% DepProbe, dislocated: 51\% BAP \& 26\% DepProbe, expl: 96\% BAP \& 91\% DepProbe, iobj: 88\% BAP \& 64\% DepProbe, nmod: 90\% BAP \& 78\% DepProbe, nsubj: 93\% BAP \& 84\% DepProbe, nummod: 96\% BAP \& 91\% DepProbe, obj: 89\% BAP \& 73\% DepProbe, obl: 88\% BAP \& 74\% DepProbe, vocative: 47\% BAP \& 31\% DepProbe. clause: acl: 89\% BAP \& 71\% DepProbe, advcl: 85\% BAP \& 64\% DepProbe, ccomp: 83\% BAP \& 51\% DepProbe, csubj: 85\% BAP \& 58\% DepProbe, xcomp: 85\% BAP \& 64\% DepProbe. modifier: advmod: 93\% BAP \& 87\% DepProbe, amod: 95\% BAP \& 90\% DepProbe, discourse: 80\% BAP \& 69\% DepProbe. function: aux: 98\% BAP \& 91\% DepProbe, case: 98\% BAP \& 97\% DepProbe, clf: 95\% BAP \& 89\% DepProbe, cop: 93\% BAP \& 63\% DepProbe, det: 97\% BAP \& 94\% DepProbe, mark: 96\% BAP \& 93\% DepProbe. coord: cc: 98\% BAP \& 97\% DepProbe, conj: 89\% BAP \& 74\% DepProbe. multi: compound: 89\% BAP \& 81\% DepProbe, fixed: 91\% BAP \& 71\% DepProbe, flat: 91\% BAP \& 81\% DepProbe. loose: list: 60\% BAP \& 53\% DepProbe, parataxis: 84\% BAP \& 60\% DepProbe. special: goeswith: 20\% BAP \& 12\% DepProbe, orphan: 47\% BAP \& 9\% DepProbe, reparandum: 22\% BAP \& 0\% DepProbe. other: dep: 72\% BAP \& 59\% DepProbe, punct: 99\% BAP \& 99\% DepProbe, ref: 80\% BAP \& 0\% DepProbe, root: 94\% BAP \& 79\% DepProbe.}
    \end{subfigure}
    \hspace{-.5em}
    \begin{subfigure}[m]{.025\textwidth}
        \includegraphics[width=\textwidth]{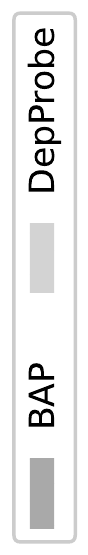}
        \vspace{.4em}
    \end{subfigure}
    \vspace{-1em}
    \caption{\textbf{Relation Accuracy of \bap{} and \depprobe{}} compared for all 13 in-language targets, grouped according to the Universal Dependencies taxonomy \citep{de-marneffe-etal-2014-universal}.}
    \label{fig:relation-errors}
\end{figure*}

\begin{figure}
    \centering
    \hspace{-1em}
    \pdftooltip{\includegraphics[width=.45\textwidth]{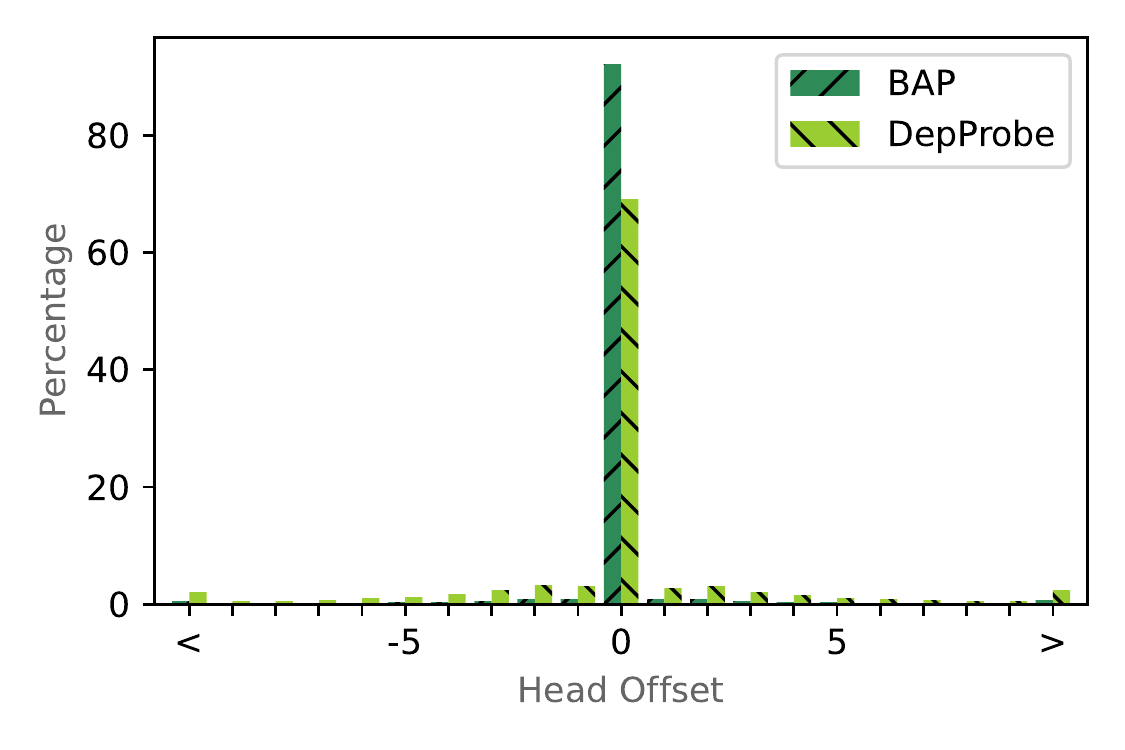}}{Screenreader Caption: 0 offset: 92\% BAP, 69\% DepProbe; 1 offset: 1\% BAP, 3\% DepProbe; -1 offset: 1\% BAP, 3\% DepProbe; 2 offset: 1\% BAP, 3\% DepProbe; -2 offset: 1\% BAP, 3\% DepProbe; 3 offset: 1\% BAP, 2\% DepProbe; -3 offset: 1\% BAP, 2\% DepProbe; 4 offset: 0.4\% BAP, 2\% DepProbe; -4 offset: 0.4\% BAP, 1\% DepProbe; 5 offset: 0.3\% BAP, 1\% DepProbe; -5 offset: 0.3\% BAP, 1\% DepProbe; greater than 5 offset: 1\% BAP, 4\% DepProbe; less than 5 offset: 1\% BAP, 4\% DepProbe.}
    \vspace{-.5em}
    \caption{\textbf{Ratio of Offsets between Gold and Predicted Heads} for \bap{} and \depprobe{} (i.e.\ 0 is correct) across all 13 targets.}
    \label{fig:head-errors}
\end{figure}

In the following analysis we investigate performance differences between the full \bap{} and \depprobe{} across all 13 targets in order to identify finer-grained limitations of the linear approach and also which kinds of dependencies benefit from full parameter tuning and non-linear decoding.

\paragraph{Edge Length} Figure \ref{fig:head-errors} shows offsets between gold and predicted head positions. The majority of heads are predicted correctly with a ratio of 92.1\% for \bap{} and 69.7\% for \depprobe{}. Both methods are less accurate in predicting long-distance edges with length 150--250, resulting in offsets of ca.\ 100 (aggregated into $<$ and $>$ in Figure \ref{fig:head-errors}). Most likely, this is due to these edges' overall sparsity in the data (only 6.7\% of edges cover a distance of more than 10 tokens) as well as their higher overall subjective difficulty. Nonetheless, \bap{} is able to capture such dependencies more accurately as shown by its lower error rates for long edges compared to those of \depprobe{}.

In addition to very distant head nodes, \bap{} also seems to recover more of the nuanced edges in the $[-5, 5]$ interval. This range is particularly impactful for downstream performance as the edges in our target treebanks have a median length of 2 (mean length 3.62 with $\sigma=$ 5.70). The structural probing loss (Equation \ref{eq:distance-loss}) and the simple linear parametrization of the probe are able to capture a large number of these edges as evidenced by overall low error rates, but lack the necessary expressivity in order to accurately capture all cases.

\paragraph{Relations} Looking at RelAcc for each category in the UD taxonomy \citep{de-marneffe-etal-2014-universal} in Figure \ref{fig:relation-errors} allows us to identify where higher parametrization and more complex decoding are required for high parsing performance. While we again observe that performance on all relations is higher for \bap{} than for \depprobe{}, a large subset of the relations is characterized by comparable or equivalent performance. These include simple punctuation (\texttt{punct}), but also the majority of function word relations such as \texttt{aux}, \texttt{case}, \texttt{clf}, \texttt{det} and \texttt{mark} as well as coordination (e.g.\ \texttt{cc}, \texttt{conj}). We attribute the high performance of \depprobe{} on these relations to the fact that the words used to express them typically stem from closed classes and consequently similar embeddings: e.g., determiners ``the/a/an'' (EN), case markers ``di/da'' (IT).

Interestingly, some relations expressed through open class words are also captured by the linear probe. These include the modifiers \texttt{advmod}, \texttt{amod} and \texttt{discourse} as well as some nominal relations such as \texttt{expl}, \texttt{nmod}, \texttt{nsubj} and \texttt{nummod}. As prior work has identified PoS information in untuned embeddings \citep{tenney2019}, the modifiers are likely benefiting from the same embedding features. The fact that \depprobe{} nonetheless identifies syntax-specific relations such as \texttt{nsubj}, and to a lesser degree \texttt{obj} and \texttt{obl}, indicates the presence of context-dependent syntactic information in addition to PoS.

The larger the set of possible words for a relation, the more difficult it is to capture with the probe. The functional \texttt{cop} (copula) relation provides an informative example: In English (and related languages), it is almost exclusively assigned to the verb ``be'' resulting in 85\% RelAcc, while in non-European languages such as Japanese it can be ascribed to a larger set which often overlaps with other relations (e.g.\ \texttt{aux}) resulting in 65\% RelAcc. \bap{} adapts to each language by tuning all parameters while \depprobe{}, using fixed embeddings, reaches competitive scores on European languages, but performs worse in non-European settings (details in Appendix \ref{sec:additional-results}).

Besides capturing larger variation in surface forms, \bap{} also appears to benefit from higher expressivity when labeling clausal relations such as \texttt{ccomp}, \texttt{csubj}. These relations are often characterized not only by surface form variation, but also by PoS variation of head/child words and overlap with other relation types (e.g.\ clausal subjects stem from verbs or adjectives), making them difficult to distinguish in untuned embeddings. Simultaneously, they often span longer edges compared to determiners or other function words.

Another relation of particular importance is \texttt{root} as it determines the direction of all edges predicted by \depprobe{}. An analysis of the 14\% RelAcc difference to \bap{} reveals that both methods most frequently confuse \texttt{root} with relations that fit the word's PoS, e.g.\ \texttt{NOUN} roots with \texttt{nsubj} or \texttt{nmod}. For the majority PoS \texttt{VERB} (70\% of all \texttt{root}), we further observe that \depprobe{} predicts twice as many \texttt{xcomp} and \texttt{parataxis} confusions compared to \bap{}, likely attributable to their \texttt{root}-similar function in subclauses. Since their distinction hinges on context, the full parser, which also tunes the contextual encoder, is better equipped to differentiate between them.

The last category in which \bap{} outperforms \depprobe{} includes rare, treebank-specific relations such as \texttt{reparandum} (reference from a corrected word to an erroneous one). Again, the larger number of tunable parameters in addition to the non-linear decoding procedure of the full parser enable it to capture more edge cases while \depprobe{}'s linear approach can only approximate a local optimum for any relations which are represented non-linearly.

\paragraph{Efficiency} When using a probe for performance prediction, it is important to consider its computational efficiency over the full parser's fine-tuning procedure. In terms of tunable parameters, \depprobe{} has 36\% fewer parameters than \dirprobe{} and three orders of magnitude fewer parameters than \bap{}. In practice, this translates to training times in the order of minutes instead of hours.

Despite its simple $\mathcal{O}(n^2)$ decoding procedure compared to dynamic programming-based graph-decoding algorithms ($\mathcal{O}(n^3)$), \depprobe{} is able to extract full dependency trees which correlate highly with downstream performance while maintaining high efficiency (Section \ref{sec:transfer-prediction}).

\section{Conclusion}\label{sec:conclusion}

With \depprobe{}, we have introduced a novel probing procedure to extract fully labeled and directed dependency trees from untuned, contextualized embeddings. Compared to prior approaches which extract structures lacking labels, edge directionality or both, our method retains a simple linear parametrization which is in fact more lightweight and does not require complex decoders (Section \ref{sec:probing}).

To the best of our knowledge, this is the first linear probe which can be used to estimate LAS from untuned embeddings. Using this property, we evaluated the predictive power of \depprobe{} on cross-lingual parsing with respect to the transfer performance of a fully fine-tuned biaffine attention parser. Across the considered 169 language pairs, \depprobe{} is surprisingly effective: Its LAS correlates significantly ($p < 0.001$) and most highly compared with unlabeled probes or competitive language feature baselines, choosing the best source treebank in 94\% of all cases (Section \ref{sec:experiments}).

Leveraging the linearity of the probe to analyze structural and relational subspaces in mBERT embeddings, we find that dependency \textit{relation} information is particularly important for parsing performance and cross-lingual transferability, compared to both tree depth and structure. \depprobe{}, which models structure and relations, is able to recover many functional and syntactic relations with competitive accuracy to the full \bap{} (Section \ref{sec:analysis}).

Finally, the substantially higher efficiency of \depprobe{} with respect to time and compute make it suitable for accurate parsing performance prediction. As contemporary performance prediction methods lack formulations for graphical tasks and handcrafted features such as lang2vec are not available in all transfer settings (e.g.\ document domains, MLM encoder choice), we see linear approaches such as \depprobe{} as a valuable alternative.

%
%
\section*{Acknowledgements}
We would like to thank the NLPnorth group for insightful discussions on this work --- in particular Elisa Bassignana and Mike Zhang. Additional thanks to ITU's High-performance Computing Cluster team. Finally, we thank the anonymous reviewers for their helpful feedback. This research is supported by the Independent Research Fund Denmark (Danmarks Frie Forskningsfond; DFF) grant number 9063-00077B.

\bibliography{anthology,references}
\bibliographystyle{acl_natbib}

\clearpage

%
%

\appendix

\section*{Appendix}\label{sec:appendix}

\section{Experimental Setup}\label{sec:experiment-setup}

\begin{table}[h!]
\centering
\resizebox{.47\textwidth}{!}{
\begin{tabular}{lllr}
\toprule
\textsc{Target} & \textsc{Lang} & \textsc{Family} & \multicolumn{1}{l}{\textsc{Size}} \\
\midrule
AR-PADT & Arabic & Afro-Asiatic & 7.6k \\
EN-EWT & English & Indo-European & 16.6k \\
EU-BDT & Basque & Basque & 9.0k \\
FI-TDT & Finnish & Uralic & 15.1k \\
HE-HTB & Hebrew & Afro-Asiatic & 6.2k \\
HI-HDTB & Hindi & Indo-European & 16.6k \\
IT-ISDT & Italian & Indo-European & 14.1k \\
JA-GSD & Japanese & Japanese & 8.1k \\
KO-GSD & Korean & Korean & 6.3k \\
RU-SynTagRus & Russian & Indo-European & 61.9k \\
SV-Talbanken & Swedish & Indo-European & 6.0k \\
TR-IMST & Turkish & Turkic & 5.6k \\
ZH-GSD & Chinese & Sino-Tibetan & 5.0k \\
\bottomrule
\end{tabular}
}
\caption{\label{tab:target-treebanks} \textbf{Target Treebanks} based on \citet{kulmizev-etal-2019-deep} with language family (\textsc{Family}) and total number of sentences (\textsc{Size}).}
\end{table}

\paragraph{Target Treebanks} Table \ref{tab:target-treebanks} lists the 13 target treebanks based on the set by \citet{kulmizev-etal-2019-deep}: AR-PADT \citep{UD_Arabic-PADT}, EN-EWT \citep{silveira-etal-2014-gold}, EU-BDT \citep{aranzabe2015automatic}, FI-TDT \citep{pyysalo-etal-2015-universal}, HE-HTB \citep{mcdonald-etal-2013-universal}, HI-HDTB \citep{palmer2009hindi}, IT-ISDT \citep{bosco2014evalita}, JA-GSD \citep{asahara-etal-2018-universal}, KO-GSD \citep{chun-etal-2018-building}, RU-SynTagRus \citep{droganova2018data}, SV-Talbanken \citep{mcdonald-etal-2013-universal}, TR-IMST \citep{sulubacak-etal-2016-universal}, ZH-GSD \citep{ud_chinese_gsd_2016}. In our experiments, we use these treebanks as provided in Universal Dependencies version 2.8 \citep{ud28}. Each method (\bap{}, \depprobe{}, \dirprobe{}) is trained on each target's respective training split and evaluated on each test split both in the in-language and cross-lingual setting without further fine-tuning. For the layer-hyperparameter of \depprobe{}, we use the development split of EN-EWT as in prior probing work \citep{hewitt2019}.

\paragraph{Implementation} \bap{} \citep{dozat2017} uses the implementation in the MaChAmp toolkit v0.2 \citep{van-der-goot-etal-2021-massive} with the default training schedule and hyperparameters. \dirprobe{} \citep{kulmizev-etal-2020-neural} is reimplemented based on the authors' algorithm description and uses their reported hyperparameters. Both it and \depprobe{} (this work) are implemented in PyTorch v1.9.0 \citep{pytorch} and use mBERT (\texttt{bert-base-multilingual-cased}) from the Transformers library v4.8.2 \citep{wolf-etal-2020-transformers}. Following prior probing work, each token which is split by mBERT into multiple subwords is mean-pooled \citep{hewitt2019}. For lang2vec \citep{littell-etal-2017-uriel}, we use its 
\texttt{syntax\_knn}, \texttt{phonology\_knn} and \texttt{inventory\_knn} features from v1.1.2. For our analyses (Section \ref{sec:analysis}), we use numpy v1.21.0 \citep{numpy}, SciPy v1.7.0 \citep{scipy} and Matplotlib v3.4.3 \citep{matplotlib}.

\paragraph{Training Details} Each model is trained on an NVIDIA A100 GPU with 40GBs of VRAM and an AMD Epyc 7662 CPU. Mean training time for \bap{} is ca.\ 2 h ($\pm$ 30 min). \dirprobe{} requires around 20 min ($\pm$ 5 min). \depprobe{} can be trained the fastest in around 15 min ($\pm$ 5 min) with the embedding forward operation consuming most of the time. The models use batches of size 64 and both probes have an early stopping patience of 3 (max. 30 epochs) on each target's dev data. All models are initialized thrice using the random seeds 41, 42 and 43.

\paragraph{Reproducibility} In order to ensure reproducibility for future work, we release the code for our methods and reimplementations in addition to token-level predictions (e.g.\ for significance testing) at \href{https://personads.me/x/acl-2022-code}{https://personads.me/x/acl-2022-code}.

\section{Additional Results}\label{sec:additional-results}

\begin{figure*}
    \centering
    \begin{subfigure}[m]{.3\textwidth}
        \pdftooltip{\includegraphics[width=\textwidth]{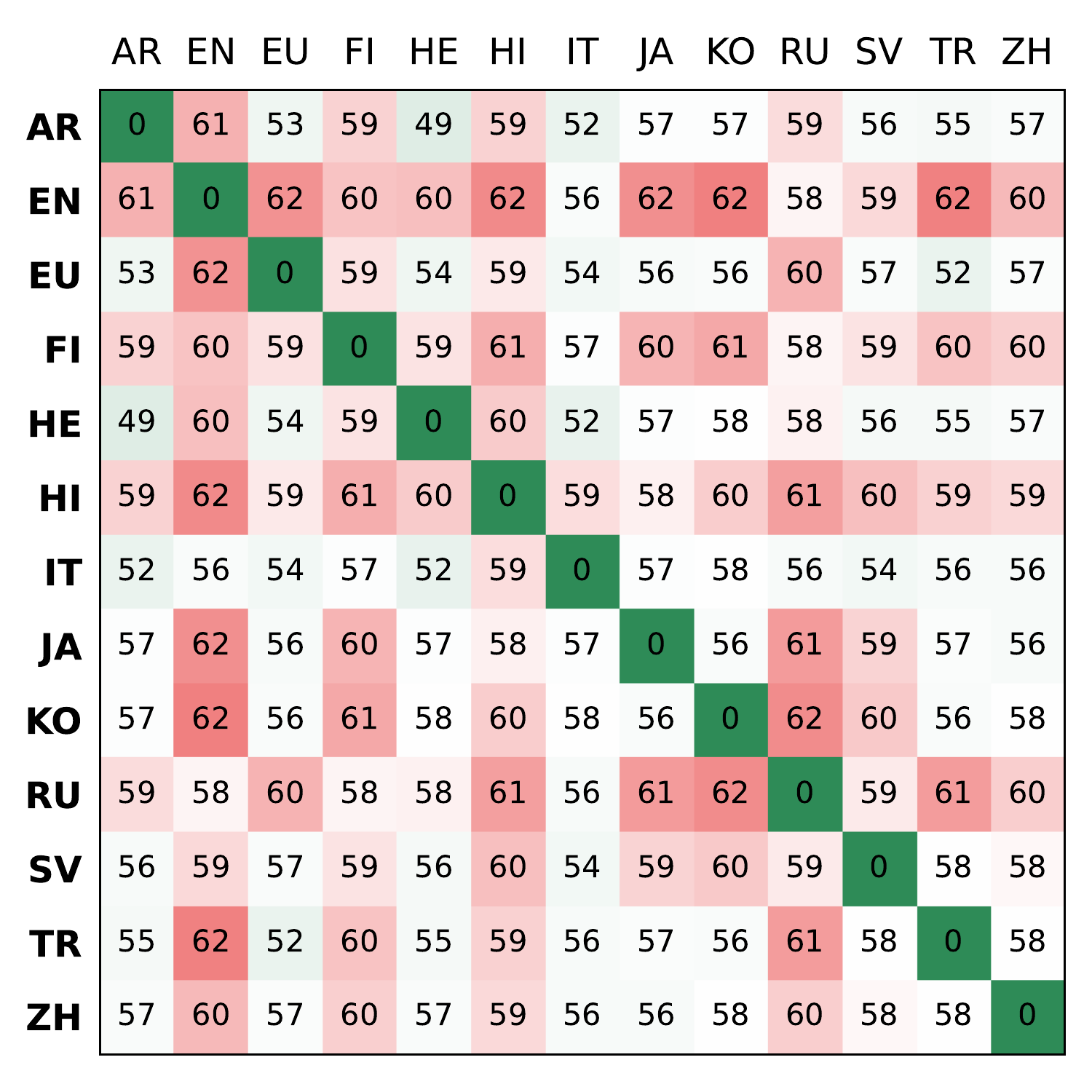}}{Screenreader Caption: AR to AR: 0, AR to EN: 61, AR to EU: 53, AR to FI: 59, AR to HE: 49, AR to HI: 59, AR to IT: 52, AR to JA: 57, AR to KO: 57, AR to RU: 59, AR to SV: 56, AR to TR: 55, AR to ZH: 57, EN to AR: 61, EN to EN: 0, EN to EU: 62, EN to FI: 60, EN to HE: 60, EN to HI: 62, EN to IT: 56, EN to JA: 62, EN to KO: 62, EN to RU: 58, EN to SV: 59, EN to TR: 62, EN to ZH: 60, EU to AR: 53, EU to EN: 62, EU to EU: 0, EU to FI: 59, EU to HE: 54, EU to HI: 59, EU to IT: 54, EU to JA: 56, EU to KO: 56, EU to RU: 60, EU to SV: 57, EU to TR: 52, EU to ZH: 57, FI to AR: 59, FI to EN: 60, FI to EU: 59, FI to FI: 0, FI to HE: 59, FI to HI: 61, FI to IT: 57, FI to JA: 60, FI to KO: 61, FI to RU: 58, FI to SV: 59, FI to TR: 60, FI to ZH: 60, HE to AR: 49, HE to EN: 60, HE to EU: 54, HE to FI: 59, HE to HE: 0, HE to HI: 60, HE to IT: 52, HE to JA: 57, HE to KO: 58, HE to RU: 58, HE to SV: 56, HE to TR: 55, HE to ZH: 57, HI to AR: 59, HI to EN: 62, HI to EU: 59, HI to FI: 61, HI to HE: 60, HI to HI: 0, HI to IT: 59, HI to JA: 58, HI to KO: 60, HI to RU: 61, HI to SV: 60, HI to TR: 59, HI to ZH: 59, IT to AR: 52, IT to EN: 56, IT to EU: 54, IT to FI: 57, IT to HE: 52, IT to HI: 59, IT to IT: 0, IT to JA: 57, IT to KO: 58, IT to RU: 56, IT to SV: 54, IT to TR: 56, IT to ZH: 56, JA to AR: 57, JA to EN: 62, JA to EU: 56, JA to FI: 60, JA to HE: 57, JA to HI: 58, JA to IT: 57, JA to JA: 0, JA to KO: 56, JA to RU: 61, JA to SV: 59, JA to TR: 57, JA to ZH: 56, KO to AR: 57, KO to EN: 62, KO to EU: 56, KO to FI: 61, KO to HE: 58, KO to HI: 60, KO to IT: 58, KO to JA: 56, KO to KO: 0, KO to RU: 62, KO to SV: 60, KO to TR: 56, KO to ZH: 58, RU to AR: 59, RU to EN: 58, RU to EU: 60, RU to FI: 58, RU to HE: 58, RU to HI: 61, RU to IT: 56, RU to JA: 61, RU to KO: 62, RU to RU: 0, RU to SV: 59, RU to TR: 61, RU to ZH: 60, SV to AR: 56, SV to EN: 59, SV to EU: 57, SV to FI: 59, SV to HE: 56, SV to HI: 60, SV to IT: 54, SV to JA: 59, SV to KO: 60, SV to RU: 59, SV to SV: 0, SV to TR: 58, SV to ZH: 58, TR to AR: 55, TR to EN: 62, TR to EU: 52, TR to FI: 60, TR to HE: 55, TR to HI: 59, TR to IT: 56, TR to JA: 57, TR to KO: 56, TR to RU: 61, TR to SV: 58, TR to TR: 0, TR to ZH: 58, ZH to AR: 57, ZH to EN: 60, ZH to EU: 57, ZH to FI: 60, ZH to HE: 57, ZH to HI: 59, ZH to IT: 56, ZH to JA: 56, ZH to KO: 58, ZH to RU: 60, ZH to SV: 58, ZH to TR: 58, ZH to ZH: 0.}
        \caption{\textsc{SSA{\footnotesize -Struct}}}
        \label{fig:ssa-structural}
    \end{subfigure}
    \begin{subfigure}[m]{.3\textwidth}
        \pdftooltip{\includegraphics[width=\textwidth]{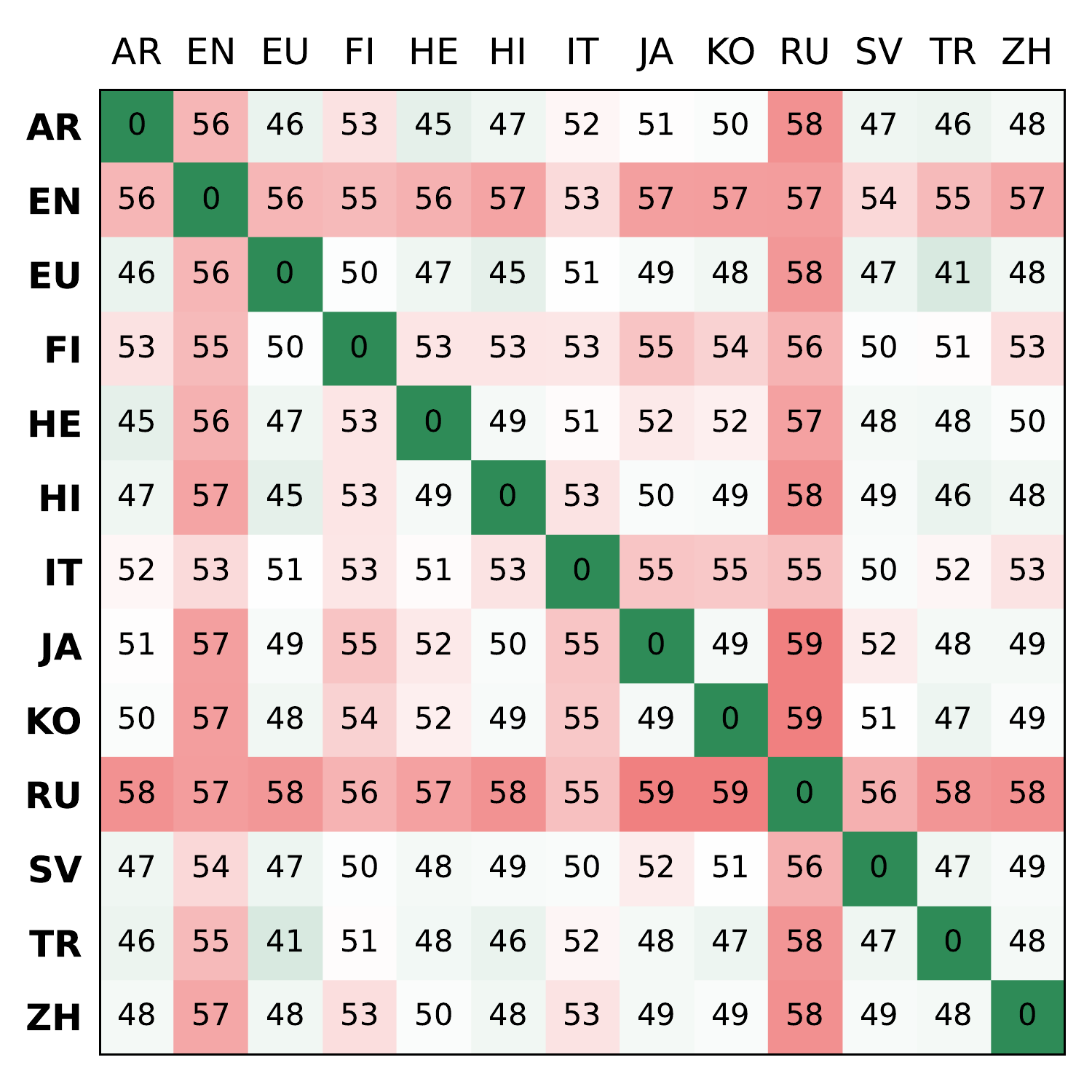}}{Screenreader Caption: AR to AR: 0, AR to EN: 56, AR to EU: 46, AR to FI: 53, AR to HE: 45, AR to HI: 47, AR to IT: 52, AR to JA: 51, AR to KO: 50, AR to RU: 58, AR to SV: 47, AR to TR: 46, AR to ZH: 48, EN to AR: 56, EN to EN: 0, EN to EU: 56, EN to FI: 55, EN to HE: 56, EN to HI: 57, EN to IT: 53, EN to JA: 57, EN to KO: 57, EN to RU: 57, EN to SV: 54, EN to TR: 55, EN to ZH: 57, EU to AR: 46, EU to EN: 56, EU to EU: 0, EU to FI: 50, EU to HE: 47, EU to HI: 45, EU to IT: 51, EU to JA: 49, EU to KO: 48, EU to RU: 58, EU to SV: 47, EU to TR: 41, EU to ZH: 48, FI to AR: 53, FI to EN: 55, FI to EU: 50, FI to FI: 0, FI to HE: 53, FI to HI: 53, FI to IT: 53, FI to JA: 55, FI to KO: 54, FI to RU: 56, FI to SV: 50, FI to TR: 51, FI to ZH: 53, HE to AR: 45, HE to EN: 56, HE to EU: 47, HE to FI: 53, HE to HE: 0, HE to HI: 49, HE to IT: 51, HE to JA: 52, HE to KO: 52, HE to RU: 57, HE to SV: 48, HE to TR: 48, HE to ZH: 50, HI to AR: 47, HI to EN: 57, HI to EU: 45, HI to FI: 53, HI to HE: 49, HI to HI: 0, HI to IT: 53, HI to JA: 50, HI to KO: 49, HI to RU: 58, HI to SV: 49, HI to TR: 46, HI to ZH: 48, IT to AR: 52, IT to EN: 53, IT to EU: 51, IT to FI: 53, IT to HE: 51, IT to HI: 53, IT to IT: 0, IT to JA: 55, IT to KO: 55, IT to RU: 55, IT to SV: 50, IT to TR: 52, IT to ZH: 53, JA to AR: 51, JA to EN: 57, JA to EU: 49, JA to FI: 55, JA to HE: 52, JA to HI: 50, JA to IT: 55, JA to JA: 0, JA to KO: 49, JA to RU: 59, JA to SV: 52, JA to TR: 48, JA to ZH: 49, KO to AR: 50, KO to EN: 57, KO to EU: 48, KO to FI: 54, KO to HE: 52, KO to HI: 49, KO to IT: 55, KO to JA: 49, KO to KO: 0, KO to RU: 59, KO to SV: 51, KO to TR: 47, KO to ZH: 49, RU to AR: 58, RU to EN: 57, RU to EU: 58, RU to FI: 56, RU to HE: 57, RU to HI: 58, RU to IT: 55, RU to JA: 59, RU to KO: 59, RU to RU: 0, RU to SV: 56, RU to TR: 58, RU to ZH: 58, SV to AR: 47, SV to EN: 54, SV to EU: 47, SV to FI: 50, SV to HE: 48, SV to HI: 49, SV to IT: 50, SV to JA: 52, SV to KO: 51, SV to RU: 56, SV to SV: 0, SV to TR: 47, SV to ZH: 49, TR to AR: 46, TR to EN: 55, TR to EU: 41, TR to FI: 51, TR to HE: 48, TR to HI: 46, TR to IT: 52, TR to JA: 48, TR to KO: 47, TR to RU: 58, TR to SV: 47, TR to TR: 0, TR to ZH: 48, ZH to AR: 48, ZH to EN: 57, ZH to EU: 48, ZH to FI: 53, ZH to HE: 50, ZH to HI: 48, ZH to IT: 53, ZH to JA: 49, ZH to KO: 49, ZH to RU: 58, ZH to SV: 49, ZH to TR: 48, ZH to ZH: 0.}
        \caption{\textsc{SSA{\footnotesize -Depth}}}
        \label{fig:ssa-depth}
    \end{subfigure}
    \begin{subfigure}[m]{.3\textwidth}
        \pdftooltip{\includegraphics[width=\textwidth]{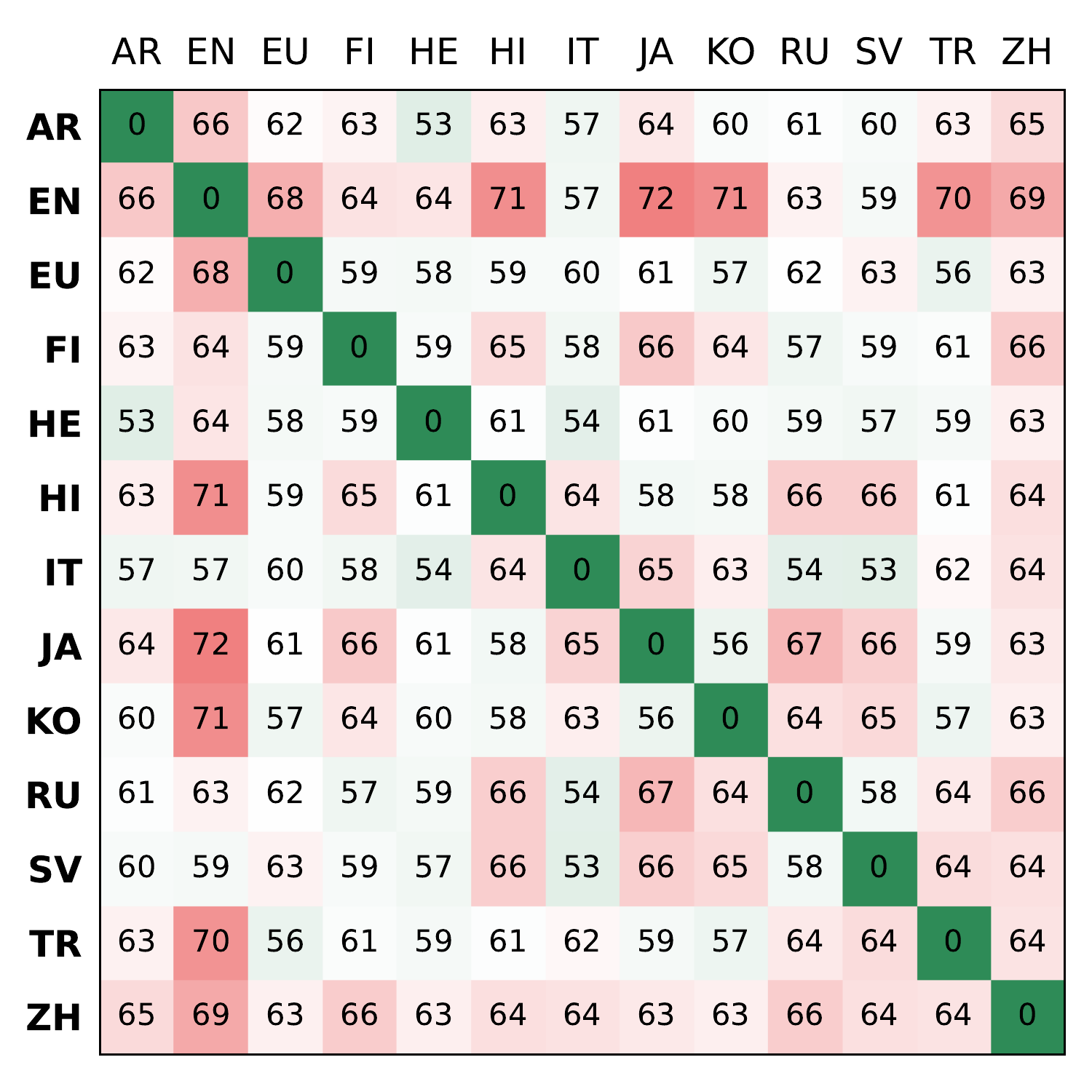}}{Screenreader Caption: AR to AR: 0, AR to EN: 66, AR to EU: 62, AR to FI: 63, AR to HE: 53, AR to HI: 63, AR to IT: 57, AR to JA: 64, AR to KO: 60, AR to RU: 61, AR to SV: 60, AR to TR: 63, AR to ZH: 65, EN to AR: 66, EN to EN: 0, EN to EU: 68, EN to FI: 64, EN to HE: 64, EN to HI: 71, EN to IT: 57, EN to JA: 72, EN to KO: 71, EN to RU: 63, EN to SV: 59, EN to TR: 70, EN to ZH: 69, EU to AR: 62, EU to EN: 68, EU to EU: 0, EU to FI: 59, EU to HE: 58, EU to HI: 59, EU to IT: 60, EU to JA: 61, EU to KO: 57, EU to RU: 62, EU to SV: 63, EU to TR: 56, EU to ZH: 63, FI to AR: 63, FI to EN: 64, FI to EU: 59, FI to FI: 0, FI to HE: 59, FI to HI: 65, FI to IT: 58, FI to JA: 66, FI to KO: 64, FI to RU: 57, FI to SV: 59, FI to TR: 61, FI to ZH: 66, HE to AR: 53, HE to EN: 64, HE to EU: 58, HE to FI: 59, HE to HE: 0, HE to HI: 61, HE to IT: 54, HE to JA: 61, HE to KO: 60, HE to RU: 59, HE to SV: 57, HE to TR: 59, HE to ZH: 63, HI to AR: 63, HI to EN: 71, HI to EU: 59, HI to FI: 65, HI to HE: 61, HI to HI: 0, HI to IT: 64, HI to JA: 58, HI to KO: 58, HI to RU: 66, HI to SV: 66, HI to TR: 61, HI to ZH: 64, IT to AR: 57, IT to EN: 57, IT to EU: 60, IT to FI: 58, IT to HE: 54, IT to HI: 64, IT to IT: 0, IT to JA: 65, IT to KO: 63, IT to RU: 54, IT to SV: 53, IT to TR: 62, IT to ZH: 64, JA to AR: 64, JA to EN: 72, JA to EU: 61, JA to FI: 66, JA to HE: 61, JA to HI: 58, JA to IT: 65, JA to JA: 0, JA to KO: 56, JA to RU: 67, JA to SV: 66, JA to TR: 59, JA to ZH: 63, KO to AR: 60, KO to EN: 71, KO to EU: 57, KO to FI: 64, KO to HE: 60, KO to HI: 58, KO to IT: 63, KO to JA: 56, KO to KO: 0, KO to RU: 64, KO to SV: 65, KO to TR: 57, KO to ZH: 63, RU to AR: 61, RU to EN: 63, RU to EU: 62, RU to FI: 57, RU to HE: 59, RU to HI: 66, RU to IT: 54, RU to JA: 67, RU to KO: 64, RU to RU: 0, RU to SV: 58, RU to TR: 64, RU to ZH: 66, SV to AR: 60, SV to EN: 59, SV to EU: 63, SV to FI: 59, SV to HE: 57, SV to HI: 66, SV to IT: 53, SV to JA: 66, SV to KO: 65, SV to RU: 58, SV to SV: 0, SV to TR: 64, SV to ZH: 64, TR to AR: 63, TR to EN: 70, TR to EU: 56, TR to FI: 61, TR to HE: 59, TR to HI: 61, TR to IT: 62, TR to JA: 59, TR to KO: 57, TR to RU: 64, TR to SV: 64, TR to TR: 0, TR to ZH: 64, ZH to AR: 65, ZH to EN: 69, ZH to EU: 63, ZH to FI: 66, ZH to HE: 63, ZH to HI: 64, ZH to IT: 64, ZH to JA: 63, ZH to KO: 63, ZH to RU: 66, ZH to SV: 64, ZH to TR: 64, ZH to ZH: 0.}
        \caption{\textsc{SSA{\footnotesize -Rel}}}
        \label{fig:ssa-relational}
    \end{subfigure}
    \caption{\textbf{SSA of Probe Transformations} in degrees across 13 target treebanks for the structural (\textsc{SSA{\footnotesize -Struct}}), depth (\textsc{SSA{\footnotesize -Depth}}) and relational probes (\textsc{SSA{\footnotesize -Rel}}).}
    \label{fig:ssa}
\end{figure*}

\paragraph{Subspace Angles} (SSA) are used in Section \ref{sec:ssa-analysis} in order to identify which types of dependency information are most relevant to final parsing performance. Figure \ref{fig:ssa} lists all cross-lingual SSAs for the structural (Figure \ref{fig:ssa-structural}), depth (Figure \ref{fig:ssa-depth}) and relational probes (Figure \ref{fig:ssa-relational}). SSA values are converted from radians to degrees $\in [0, 90]$ for improved readability. Correlation in Table \ref{tab:ssa-correlation} is calculated based on negative SSA \citep{chi2020}.

\begin{figure*}
    \begin{subfigure}[m]{.945\textwidth}
        \pdftooltip{\includegraphics[width=\textwidth]{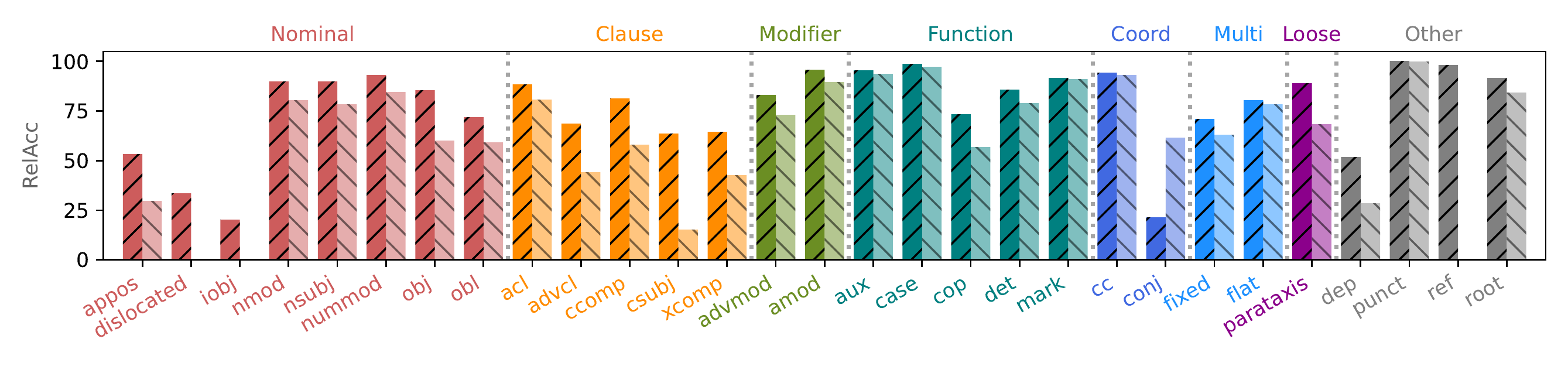}}{Screenreader Caption: nominal: appos: 53\% BAP \& 29\% DepProbe, dislocated: 33\% BAP \& 0\% DepProbe, iobj: 20\% BAP, 0\% DepProbe, nmod: 89\% BAP, 80\% DepProbe, nsubj: 89\% BAP, 78\% DepProbe, nummod: 92\% BAP, 84\% DepProbe, obj: 85\% BAP, 59\% DepProbe, obl: 71\% BAP, 58\% DepProbe. clause: acl: 88\% BAP \& 80\% DepProbe, advcl: 68\% BAP \& 43\% DepProbe, ccomp: 81\% BAP \& 57\% DepProbe, csubj: 63\% BAP \& 15\% DepProbe, xcomp: 64\% BAP \& 42\% DepProbe. modifier: advmod: 83\% BAP \& 73\% DepProbe, amod: 95\% BAP \& 89\% DepProbe. function: aux: 95\% BAP \& 93\% DepProbe, case: 98\% BAP \& 97\% DepProbe, cop: 73\% BAP \& 56\% DepProbe, det: 85\% BAP \& 78\% DepProbe, mark: 91\% BAP \& 90\% DepProbe. coord: cc: 94\% BAP \& 92\% DepProbe, conj: 21\% BAP \& 61\% DepProbe. multi: fixed: 70\% BAP \& 62\% DepProbe, flat: 80\% BAP \& 78\% DepProbe. loose: parataxis: 89\% BAP \& 68\% DepProbe. other: dep: 51\% BAP \& 28\% DepProbe, punct: 100\% BAP \& 99\% DepProbe, ref: 98\% BAP \& 0\% DepProbe, root: 91\% BAP \& 84\% DepProbe.}
    \end{subfigure}
    \hspace{-.7em}
    \begin{subfigure}[m]{.025\textwidth}
        \includegraphics[width=\textwidth]{img/all-targets-correct-rel-legend.pdf}
        \vspace{.2em}
    \end{subfigure}
    \vspace{-1em}
    \caption{\textbf{RelAcc of \bap{} and \depprobe{} on AR-PADT (Test)} grouped according to UD taxonomy.}
    \label{fig:relation-errors-ar}
\end{figure*}

\begin{figure*}
    \begin{subfigure}[m]{.945\textwidth}
        \pdftooltip{\includegraphics[width=\textwidth]{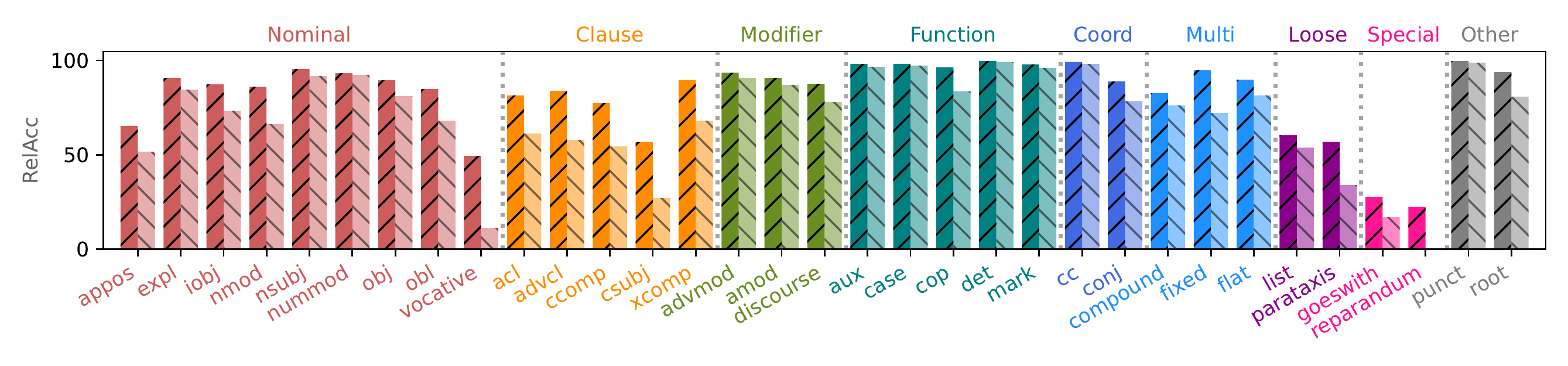}}{Screenreader Caption: nominal: appos: 65\% BAP \& 51\% DepProbe, expl: 90\% BAP \& 84\% DepProbe, iobj: 86\% BAP \& 73\% DepProbe, nmod: 85\% BAP \& 66\% DepProbe, nsubj: 95\% BAP \& 91\% DepProbe, nummod: 92\% BAP \& 92\% DepProbe, obj: 89\% BAP \& 80\% DepProbe, obl: 84\% BAP \& 68\% DepProbe, vocative: 49\% BAP \& 11\% DepProbe. clause: acl: 81\% BAP \& 61\% DepProbe, advcl: 83\% BAP \& 57\% DepProbe, ccomp: 77\% BAP \& 54\% DepProbe, csubj: 56\% BAP \& 27\% DepProbe, xcomp: 89\% BAP \& 67\% DepProbe. modifier: advmod: 93\% BAP \& 90\% DepProbe, amod: 90\% BAP \& 86\% DepProbe, discourse: 87\% BAP \& 77\% DepProbe. function: aux: 97\% BAP \& 96\% DepProbe, case: 98\% BAP \& 97\% DepProbe, cop: 96\% BAP \& 83\% DepProbe, det: 99\% BAP \& 98\% DepProbe, mark: 97\% BAP \& 95\% DepProbe. coord: cc: 98\% BAP \& 97\% DepProbe, conj: 88\% BAP \& 78\% DepProbe. multi: compound: 82\% BAP \& 76\% DepProbe, fixed: 94\% BAP \& 71\% DepProbe, flat: 89\% BAP \& 81\% DepProbe. loose: list: 60\% BAP \& 53\% DepProbe, parataxis: 56\% BAP \& 33\% DepProbe. special: goeswith: 27\% BAP \& 16\% DepProbe, reparandum: 22\% BAP \& 0\% DepProbe. other: punct: 99\% BAP \& 98\% DepProbe, root: 93\% BAP \& 80\% DepProbe.}
    \end{subfigure}
    \hspace{-.7em}
    \begin{subfigure}[m]{.025\textwidth}
        \includegraphics[width=\textwidth]{img/all-targets-correct-rel-legend.pdf}
        \vspace{.2em}
    \end{subfigure}
    \vspace{-1em}
    \caption{\textbf{RelAcc of \bap{} and \depprobe{} on EN-EWT (Test)} grouped according to UD taxonomy.}
    \label{fig:relation-errors-en}
\end{figure*}

\begin{figure*}
    \begin{subfigure}[m]{.945\textwidth}
        \pdftooltip{\includegraphics[width=\textwidth]{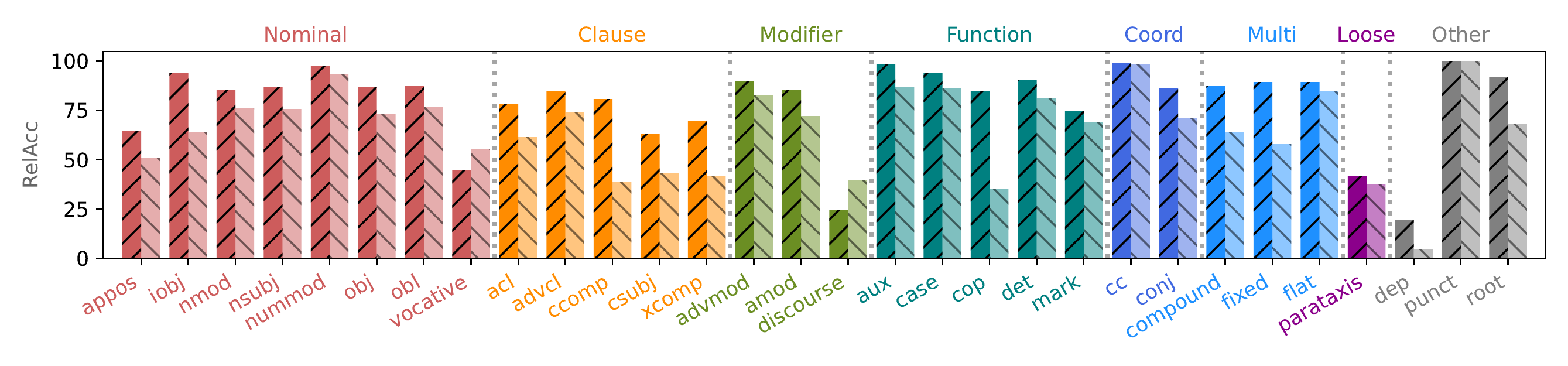}}{Screenreader Caption: nominal: appos: 64\% BAP \& 50\% DepProbe, iobj: 94\% BAP \& 64\% DepProbe, nmod: 85\% BAP \& 76\% DepProbe, nsubj: 86\% BAP \& 75\% DepProbe, nummod: 97\% BAP \& 93\% DepProbe, obj: 86\% BAP \& 73\% DepProbe, obl: 87\% BAP \& 76\% DepProbe, vocative: 44\% BAP \& 55\% DepProbe. clause: acl: 78\% BAP \& 61\% DepProbe, advcl: 84\% BAP \& 73\% DepProbe, ccomp: 80\% BAP \& 38\% DepProbe, csubj: 62\% BAP \& 42\% DepProbe, xcomp: 69\% BAP \& 41\% DepProbe. modifier: advmod: 89\% BAP \& 82\% DepProbe, amod: 85\% BAP \& 71\% DepProbe, discourse: 24\% BAP \& 39\% DepProbe. function: aux: 98\% BAP \& 86\% DepProbe, case: 93\% BAP \& 86\% DepProbe, cop: 84\% BAP \& 35\% DepProbe, det: 90\% BAP \& 80\% DepProbe, mark: 74\% BAP \& 68\% DepProbe. coord: cc: 98\% BAP \& 98\% DepProbe, conj: 86\% BAP \& 71\% DepProbe. multi: compound: 87\% BAP \& 64\% DepProbe, fixed: 89\% BAP \& 57\% DepProbe, flat: 89\% BAP \& 84\% DepProbe. loose: parataxis: 41\% BAP \& 37\% DepProbe. other: dep: 19\% BAP \& 4\% DepProbe, punct: 99\% BAP \& 99\% DepProbe, root: 91\% BAP \& 67\% DepProbe.}
    \end{subfigure}
    \hspace{-.7em}
    \begin{subfigure}[m]{.025\textwidth}
        \includegraphics[width=\textwidth]{img/all-targets-correct-rel-legend.pdf}
        \vspace{.2em}
    \end{subfigure}
    \vspace{-1em}
    \caption{\textbf{RelAcc of \bap{} and \depprobe{} on EU-BDT (Test)} grouped according to UD taxonomy.}
    \label{fig:relation-errors-eu}
\end{figure*}

\begin{figure*}
    \begin{subfigure}[m]{.945\textwidth}
        \pdftooltip{\includegraphics[width=\textwidth]{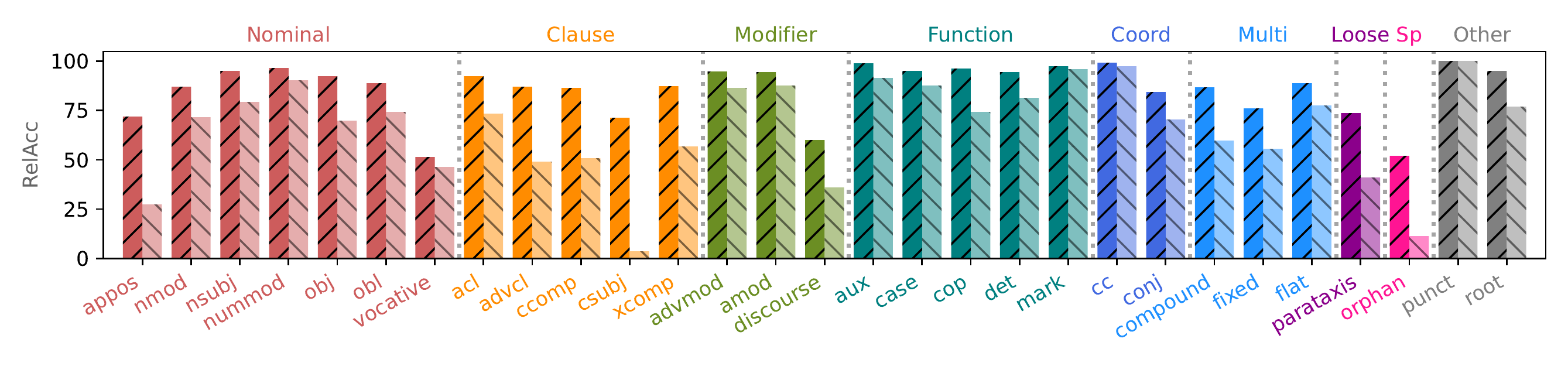}}{Screenreader Caption: nominal: appos: 71\% BAP \& 27\% DepProbe, nmod: 86\% BAP \& 71\% DepProbe, nsubj: 94\% BAP \& 79\% DepProbe, nummod: 96\% BAP \& 90\% DepProbe, obj: 92\% BAP \& 69\% DepProbe, obl: 88\% BAP \& 74\% DepProbe, vocative: 51\% BAP \& 46\% DepProbe. clause: acl: 92\% BAP \& 73\% DepProbe, advcl: 86\% BAP \& 48\% DepProbe, ccomp: 86\% BAP \& 50\% DepProbe, csubj: 71\% BAP \& 3\% DepProbe, xcomp: 87\% BAP \& 56\% DepProbe. modifier: advmod: 94\% BAP \& 86\% DepProbe, amod: 94\% BAP \& 87\% DepProbe, discourse: 60\% BAP \& 36\% DepProbe. function: aux: 98\% BAP \& 91\% DepProbe, case: 95\% BAP \& 87\% DepProbe, cop: 96\% BAP \& 74\% DepProbe, det: 94\% BAP \& 81\% DepProbe, mark: 97\% BAP \& 95\% DepProbe. coord: cc: 99\% BAP \& 97\% DepProbe, conj: 84\% BAP \& 70\% DepProbe. multi: compound: 86\% BAP \& 59\% DepProbe, fixed: 75\% BAP \& 55\% DepProbe, flat: 88\% BAP \& 77\% DepProbe. loose: parataxis: 73\% BAP \& 40\% DepProbe. special: orphan: 51\% BAP \& 11\% DepProbe. other: punct: 99\% BAP \& 99\% DepProbe, root: 94\% BAP \& 76\% DepProbe.}
    \end{subfigure}
    \hspace{-.7em}
    \begin{subfigure}[m]{.025\textwidth}
        \includegraphics[width=\textwidth]{img/all-targets-correct-rel-legend.pdf}
        \vspace{.2em}
    \end{subfigure}
    \vspace{-1em}
    \caption{\textbf{RelAcc of \bap{} and \depprobe{} on FI-TDT (Test)} grouped according to UD taxonomy.}
    \label{fig:relation-errors-fi}
\end{figure*}

\begin{figure*}
    \begin{subfigure}[m]{.945\textwidth}
        \pdftooltip{\includegraphics[width=\textwidth]{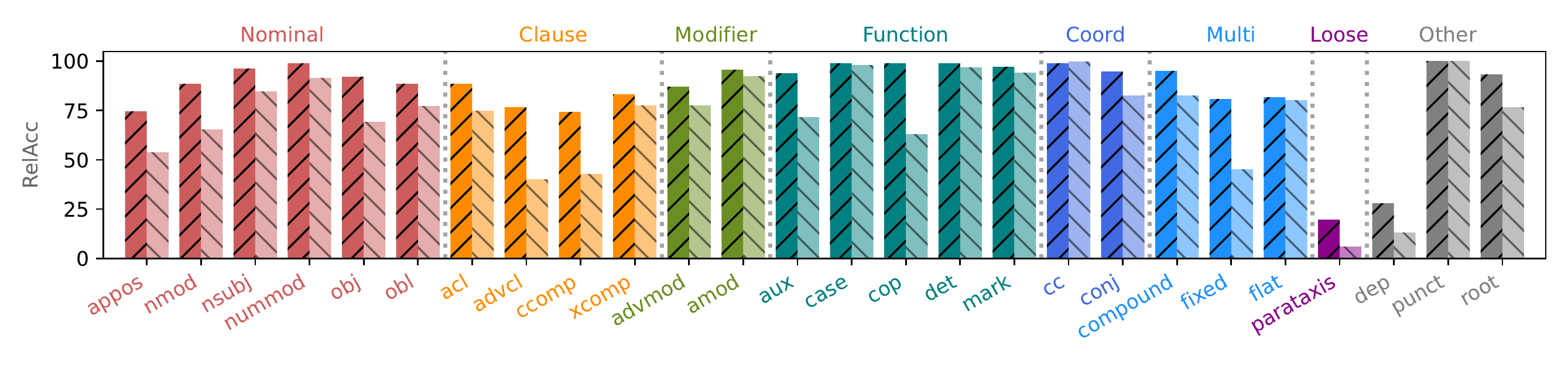}}{Screenreader Caption: nominal: appos: 74\% BAP \& 53\% DepProbe, nmod: 88\% BAP \& 65\% DepProbe, nsubj: 96\% BAP \& 84\% DepProbe, nummod: 98\% BAP \& 91\% DepProbe, obj: 92\% BAP \& 69\% DepProbe, obl: 88\% BAP \& 77\% DepProbe. clause: acl: 88\% BAP \& 74\% DepProbe, advcl: 76\% BAP \& 40\% DepProbe, ccomp: 74\% BAP \& 42\% DepProbe, xcomp: 83\% BAP \& 77\% DepProbe. modifier: advmod: 86\% BAP \& 77\% DepProbe, amod: 95\% BAP \& 92\% DepProbe. function: aux: 93\% BAP \& 71\% DepProbe, case: 98\% BAP \& 97\% DepProbe, cop: 99\% BAP \& 62\% DepProbe, det: 98\% BAP \& 96\% DepProbe, mark: 96\% BAP \& 94\% DepProbe. coord: cc: 98\% BAP \& 99\% DepProbe, conj: 94\% BAP \& 82\% DepProbe. multi: compound: 94\% BAP \& 82\% DepProbe, fixed: 80\% BAP \& 45\% DepProbe, flat: 81\% BAP \& 80\% DepProbe. loose: parataxis: 19\% BAP \& 5\% DepProbe. other: dep: 27\% BAP \& 12\% DepProbe, punct: 100\% BAP \& 99\% DepProbe, root: 93\% BAP \& 76\% DepProbe.}
    \end{subfigure}
    \hspace{-.7em}
    \begin{subfigure}[m]{.025\textwidth}
        \includegraphics[width=\textwidth]{img/all-targets-correct-rel-legend.pdf}
        \vspace{.2em}
    \end{subfigure}
    \vspace{-1em}
    \caption{\textbf{RelAcc of \bap{} and \depprobe{} on HE-HTB (Test)} grouped according to UD taxonomy.}
    \label{fig:relation-errors-he}
\end{figure*}

\begin{figure*}
    \begin{subfigure}[m]{.945\textwidth}
        \pdftooltip{\includegraphics[width=\textwidth]{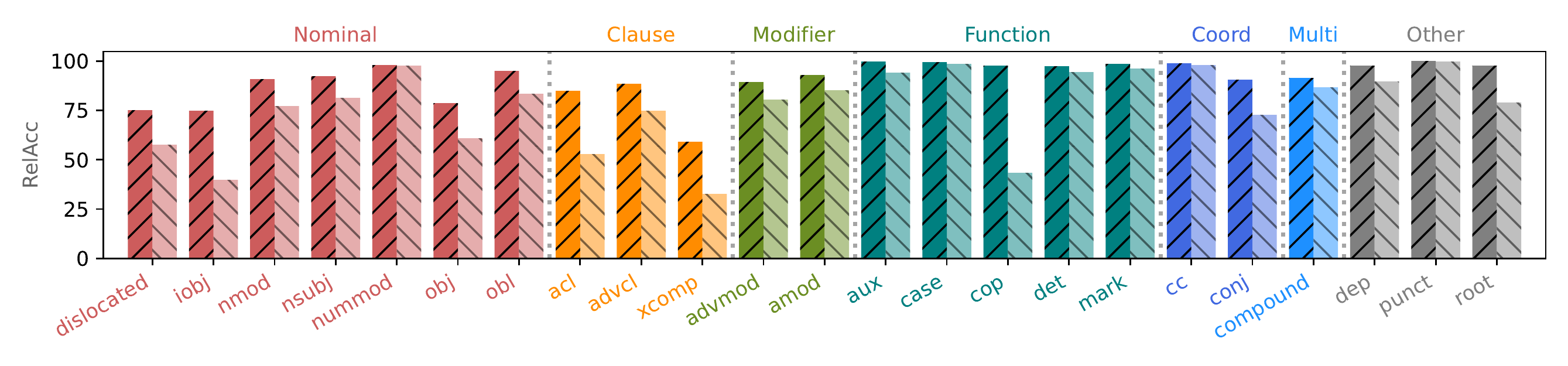}}{Screenreader Caption: nominal: dislocated: 75\% BAP \& 57\% DepProbe, iobj: 74\% BAP \& 39\% DepProbe, nmod: 90\% BAP \& 77\% DepProbe, nsubj: 92\% BAP \& 81\% DepProbe, nummod: 98\% BAP \& 97\% DepProbe, obj: 78\% BAP \& 60\% DepProbe, obl: 94\% BAP \& 83\% DepProbe. clause: acl: 84\% BAP \& 52\% DepProbe, advcl: 88\% BAP \& 74\% DepProbe, xcomp: 58\% BAP \& 32\% DepProbe. modifier: advmod: 89\% BAP \& 80\% DepProbe, amod: 92\% BAP \& 85\% DepProbe. function: aux: 99\% BAP \& 93\% DepProbe, case: 99\% BAP \& 98\% DepProbe, cop: 97\% BAP \& 43\% DepProbe, det: 97\% BAP \& 94\% DepProbe, mark: 98\% BAP \& 96\% DepProbe. coord: cc: 98\% BAP \& 97\% DepProbe, conj: 90\% BAP \& 72\% DepProbe. multi: compound: 91\% BAP \& 86\% DepProbe. other: dep: 97\% BAP \& 89\% DepProbe, punct: 99\% BAP \& 99\% DepProbe, root: 97\% BAP \& 78\% DepProbe.}
    \end{subfigure}
    \hspace{-.7em}
    \begin{subfigure}[m]{.025\textwidth}
        \includegraphics[width=\textwidth]{img/all-targets-correct-rel-legend.pdf}
        \vspace{.2em}
    \end{subfigure}
    \vspace{-1em}
    \caption{\textbf{RelAcc of \bap{} and \depprobe{} on HI-HDTB (Test)} grouped according to UD taxonomy.}
    \label{fig:relation-errors-hi}
\end{figure*}

\begin{figure*}
    \begin{subfigure}[m]{.945\textwidth}
        \pdftooltip{\includegraphics[width=\textwidth]{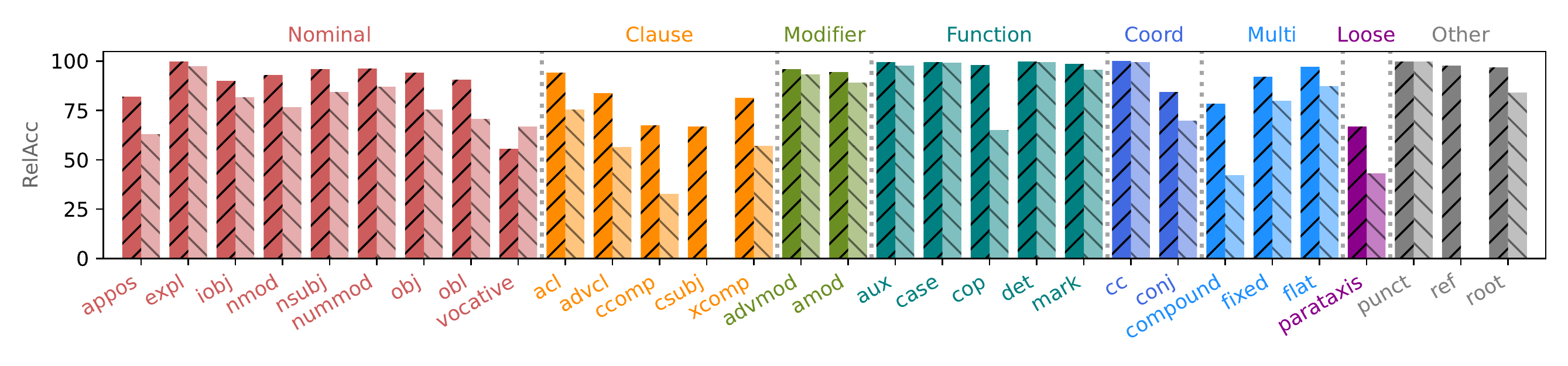}}{Screenreader Caption: nominal: appos: 81\% BAP \& 63\% DepProbe, expl: 99\% BAP \& 97\% DepProbe, iobj: 90\% BAP \& 81\% DepProbe, nmod: 92\% BAP \& 76\% DepProbe, nsubj: 95\% BAP \& 84\% DepProbe, nummod: 96\% BAP \& 86\% DepProbe, obj: 94\% BAP \& 75\% DepProbe, obl: 90\% BAP \& 70\% DepProbe, vocative: 55\% BAP \& 66\% DepProbe. clause: acl: 94\% BAP \& 75\% DepProbe, advcl: 83\% BAP \& 56\% DepProbe, ccomp: 67\% BAP \& 32\% DepProbe, csubj: 66\% BAP \& 0\% DepProbe, xcomp: 81\% BAP \& 56\% DepProbe. modifier: advmod: 95\% BAP \& 93\% DepProbe, amod: 94\% BAP \& 89\% DepProbe. function: aux: 99\% BAP \& 97\% DepProbe, case: 99\% BAP \& 99\% DepProbe, cop: 98\% BAP \& 65\% DepProbe, det: 99\% BAP \& 99\% DepProbe, mark: 98\% BAP \& 95\% DepProbe. coord: cc: 100\% BAP \& 99\% DepProbe, conj: 84\% BAP \& 69\% DepProbe. multi: compound: 78\% BAP \& 42\% DepProbe, fixed: 92\% BAP \& 79\% DepProbe, flat: 97\% BAP \& 87\% DepProbe. loose: parataxis: 66\% BAP \& 42\% DepProbe. other: punct: 99\% BAP \& 99\% DepProbe, ref: 97\% BAP \& 0\% DepProbe, root: 96\% BAP \& 84\% DepProbe.}
    \end{subfigure}
    \hspace{-.7em}
    \begin{subfigure}[m]{.025\textwidth}
        \includegraphics[width=\textwidth]{img/all-targets-correct-rel-legend.pdf}
        \vspace{.2em}
    \end{subfigure}
    \vspace{-1em}
    \caption{\textbf{RelAcc of \bap{} and \depprobe{} on IT-ISDT (Test)} grouped according to UD taxonomy.}
    \label{fig:relation-errors-it}
\end{figure*}

\begin{figure*}
    \begin{subfigure}[m]{.945\textwidth}
        \pdftooltip{\includegraphics[width=\textwidth]{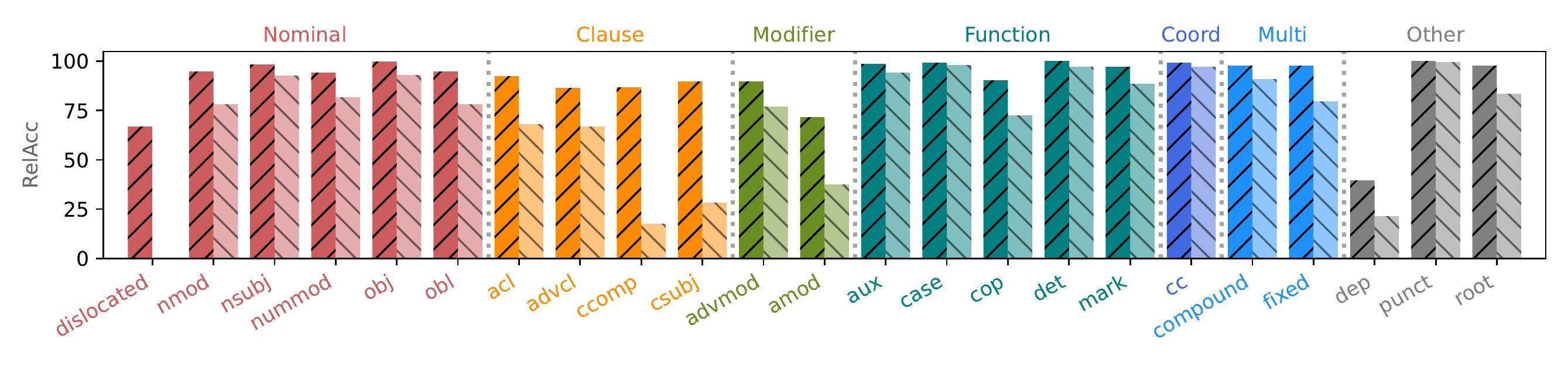}}{Screenreader Caption: nominal: dislocated: 66\% BAP \& 0\% DepProbe, nmod: 94\% BAP \& 78\% DepProbe, nsubj: 98\% BAP \& 92\% DepProbe, nummod: 94\% BAP \& 81\% DepProbe, obj: 99\% BAP \& 93\% DepProbe, obl: 94\% BAP \& 78\% DepProbe. clause: acl: 92\% BAP \& 67\% DepProbe, advcl: 86\% BAP \& 66\% DepProbe, ccomp: 86\% BAP \& 17\% DepProbe, csubj: 89\% BAP \& 28\% DepProbe. modifier: advmod: 89\% BAP \& 76\% DepProbe, amod: 71\% BAP \& 37\% DepProbe. function: aux: 98\% BAP \& 94\% DepProbe, case: 99\% BAP \& 98\% DepProbe, cop: 90\% BAP \& 72\% DepProbe, det: 100\% BAP \& 97\% DepProbe, mark: 97\% BAP \& 88\% DepProbe. coord: cc: 99\% BAP \& 96\% DepProbe. multi: compound: 97\% BAP \& 90\% DepProbe, fixed: 97\% BAP \& 79\% DepProbe. other: dep: 39\% BAP \& 21\% DepProbe, punct: 100\% BAP \& 99\% DepProbe, root: 97\% BAP \& 83\% DepProbe.}
    \end{subfigure}
    \hspace{-.7em}
    \begin{subfigure}[m]{.025\textwidth}
        \includegraphics[width=\textwidth]{img/all-targets-correct-rel-legend.pdf}
        \vspace{.2em}
    \end{subfigure}
    \vspace{-1em}
    \caption{\textbf{RelAcc of \bap{} and \depprobe{} on JA-GSD (Test)} grouped according to UD taxonomy.}
    \label{fig:relation-errors-ja}
\end{figure*}

\begin{figure*}
    \begin{subfigure}[m]{.945\textwidth}
        \pdftooltip{\includegraphics[width=\textwidth]{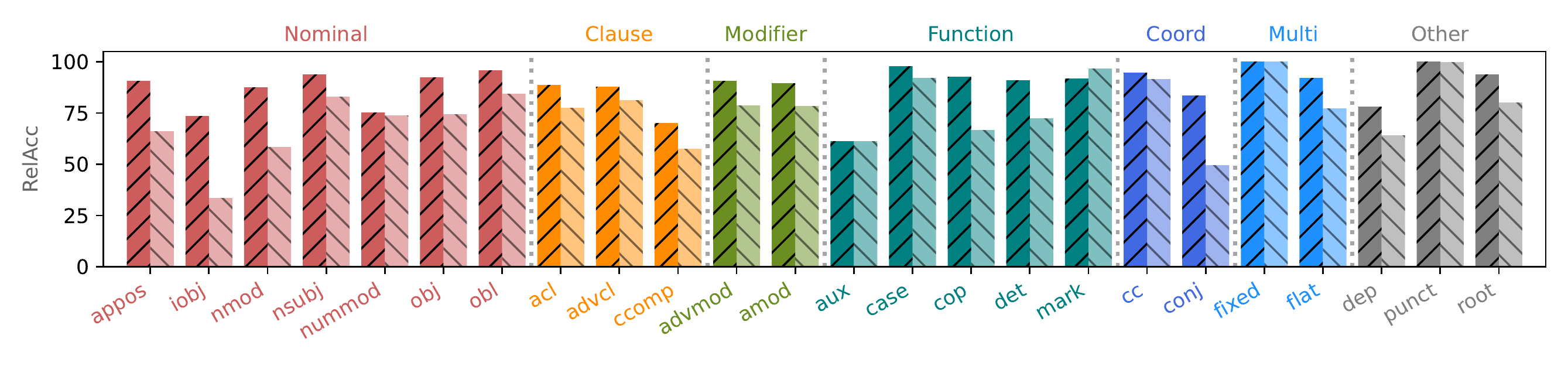}}{Screenreader Caption: nominal: appos: 90\% BAP \& 65\% DepProbe, iobj: 73\% BAP \& 33\% DepProbe, nmod: 87\% BAP \& 58\% DepProbe, nsubj: 93\% BAP \& 82\% DepProbe, nummod: 75\% BAP \& 73\% DepProbe, obj: 92\% BAP \& 74\% DepProbe, obl: 95\% BAP \& 84\% DepProbe. clause: acl: 88\% BAP \& 77\% DepProbe, advcl: 87\% BAP \& 81\% DepProbe, ccomp: 69\% BAP \& 57\% DepProbe. modifier: advmod: 90\% BAP \& 78\% DepProbe, amod: 89\% BAP \& 78\% DepProbe. function: aux: 61\% BAP \& 61\% DepProbe, case: 97\% BAP \& 91\% DepProbe, cop: 92\% BAP \& 66\% DepProbe, det: 90\% BAP \& 72\% DepProbe, mark: 91\% BAP \& 96\% DepProbe. coord: cc: 94\% BAP \& 91\% DepProbe, conj: 83\% BAP \& 49\% DepProbe. multi: fixed: 100\% BAP \& 100\% DepProbe, flat: 92\% BAP \& 77\% DepProbe. other: dep: 77\% BAP \& 64\% DepProbe, punct: 99\% BAP \& 99\% DepProbe, root: 93\% BAP \& 79\% DepProbe.}
    \end{subfigure}
    \hspace{-.7em}
    \begin{subfigure}[m]{.025\textwidth}
        \includegraphics[width=\textwidth]{img/all-targets-correct-rel-legend.pdf}
        \vspace{.2em}
    \end{subfigure}
    \vspace{-1em}
    \caption{\textbf{RelAcc of \bap{} and \depprobe{} on KO-GSD (Test)} grouped according to UD taxonomy.}
    \label{fig:relation-errors-ko}
\end{figure*}

\begin{figure*}
    \begin{subfigure}[m]{.945\textwidth}
        \pdftooltip{\includegraphics[width=\textwidth]{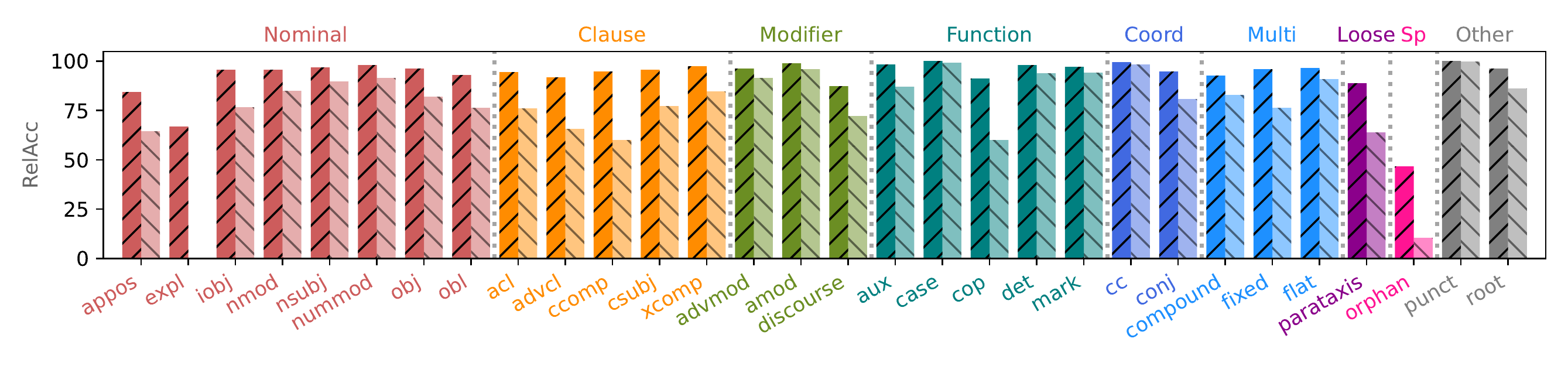}}{Screenreader Caption: nominal: appos: 84\% BAP \& 64\% DepProbe, expl: 66\% BAP \& 0\% DepProbe, iobj: 95\% BAP \& 76\% DepProbe, nmod: 95\% BAP \& 84\% DepProbe, nsubj: 96\% BAP \& 89\% DepProbe, nummod: 97\% BAP \& 91\% DepProbe, obj: 96\% BAP \& 82\% DepProbe, obl: 92\% BAP \& 76\% DepProbe. clause: acl: 94\% BAP \& 75\% DepProbe, advcl: 91\% BAP \& 65\% DepProbe, ccomp: 94\% BAP \& 60\% DepProbe, csubj: 95\% BAP \& 77\% DepProbe, xcomp: 97\% BAP \& 84\% DepProbe. modifier: advmod: 96\% BAP \& 91\% DepProbe, amod: 98\% BAP \& 95\% DepProbe, discourse: 87\% BAP \& 72\% DepProbe. function: aux: 98\% BAP \& 86\% DepProbe, case: 99\% BAP \& 99\% DepProbe, cop: 91\% BAP \& 59\% DepProbe, det: 97\% BAP \& 93\% DepProbe, mark: 96\% BAP \& 93\% DepProbe. coord: cc: 99\% BAP \& 98\% DepProbe, conj: 94\% BAP \& 80\% DepProbe. multi: compound: 92\% BAP \& 82\% DepProbe, fixed: 95\% BAP \& 76\% DepProbe, flat: 96\% BAP \& 90\% DepProbe. loose: parataxis: 88\% BAP \& 63\% DepProbe. special: orphan: 46\% BAP \& 10\% DepProbe. other: punct: 99\% BAP \& 99\% DepProbe, root: 96\% BAP \& 85\% DepProbe.}
    \end{subfigure}
    \hspace{-.7em}
    \begin{subfigure}[m]{.025\textwidth}
        \includegraphics[width=\textwidth]{img/all-targets-correct-rel-legend.pdf}
        \vspace{.2em}
    \end{subfigure}
    \vspace{-1em}
    \caption{\textbf{RelAcc of \bap{} and \depprobe{} on RU-SynTagRus (Test)} grouped according to UD taxonomy.}
    \label{fig:relation-errors-ru}
\end{figure*}

\begin{figure*}
    \begin{subfigure}[m]{.945\textwidth}
        \pdftooltip{\includegraphics[width=\textwidth]{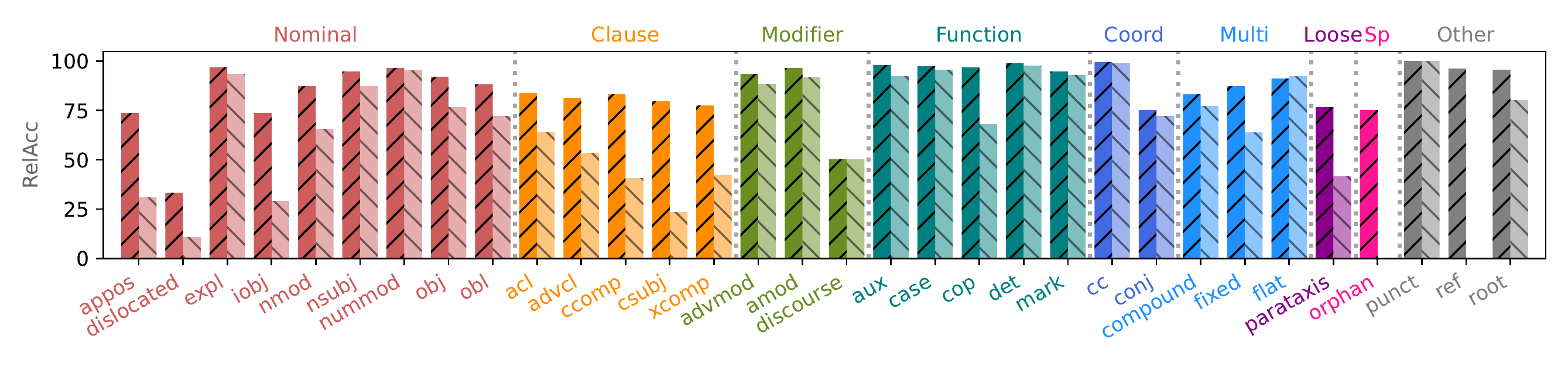}}{Screenreader Caption: nominal: appos: 73\% BAP \& 30\% DepProbe, dislocated: 33\% BAP \& 10\% DepProbe, expl: 96\% BAP \& 93\% DepProbe, iobj: 73\% BAP \& 29\% DepProbe, nmod: 87\% BAP \& 65\% DepProbe, nsubj: 94\% BAP \& 87\% DepProbe, nummod: 96\% BAP \& 95\% DepProbe, obj: 92\% BAP \& 76\% DepProbe, obl: 88\% BAP \& 72\% DepProbe. clause: acl: 83\% BAP \& 64\% DepProbe, advcl: 81\% BAP \& 53\% DepProbe, ccomp: 83\% BAP \& 40\% DepProbe, csubj: 79\% BAP \& 23\% DepProbe, xcomp: 77\% BAP \& 42\% DepProbe. modifier: advmod: 93\% BAP \& 88\% DepProbe, amod: 96\% BAP \& 91\% DepProbe, discourse: 50\% BAP \& 50\% DepProbe. function: aux: 98\% BAP \& 92\% DepProbe, case: 97\% BAP \& 95\% DepProbe, cop: 96\% BAP \& 67\% DepProbe, det: 98\% BAP \& 97\% DepProbe, mark: 94\% BAP \& 92\% DepProbe. coord: cc: 99\% BAP \& 98\% DepProbe, conj: 75\% BAP \& 72\% DepProbe. multi: compound: 83\% BAP \& 77\% DepProbe, fixed: 87\% BAP \& 63\% DepProbe, flat: 91\% BAP \& 92\% DepProbe. loose: parataxis: 76\% BAP \& 41\% DepProbe. special: orphan: 75\% BAP \& 0\% DepProbe. other: punct: 99\% BAP \& 100\% DepProbe, ref: 96\% BAP \& 0\% DepProbe, root: 95\% BAP \& 80\% DepProbe.}
    \end{subfigure}
    \hspace{-.7em}
    \begin{subfigure}[m]{.025\textwidth}
        \includegraphics[width=\textwidth]{img/all-targets-correct-rel-legend.pdf}
        \vspace{.2em}
    \end{subfigure}
    \vspace{-1em}
    \caption{\textbf{RelAcc of \bap{} and \depprobe{} on SV-Talbanken (Test)} grouped according to UD taxonomy.}
    \label{fig:relation-errors-sv}
\end{figure*}

\begin{figure*}
    \begin{subfigure}[m]{.945\textwidth}
        \pdftooltip{\includegraphics[width=\textwidth]{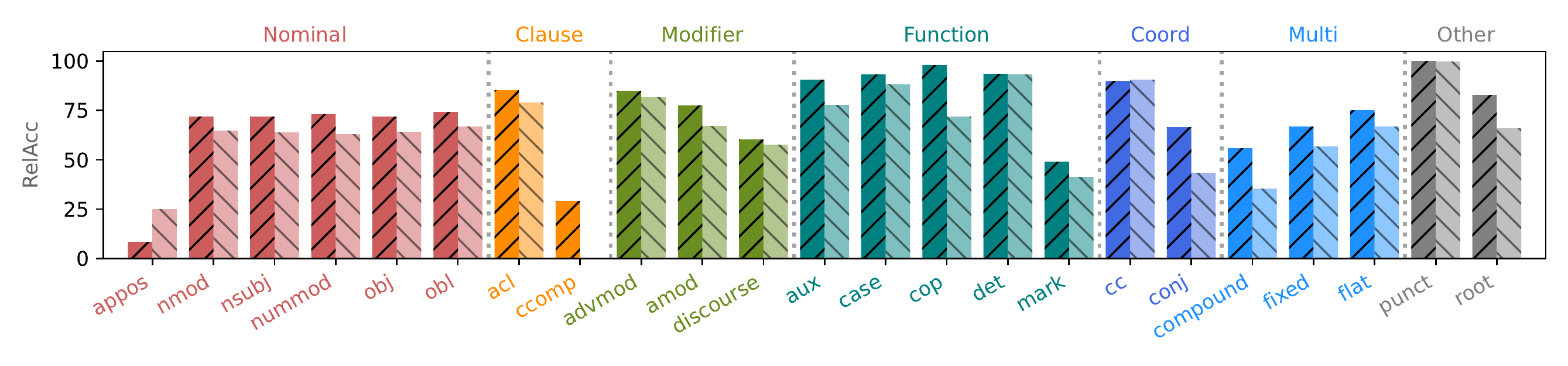}}{Screenreader Caption: nominal: appos: 8\% BAP \& 25\% DepProbe, nmod: 71\% BAP \& 64\% DepProbe, nsubj: 71\% BAP \& 63\% DepProbe, nummod: 73\% BAP \& 62\% DepProbe, obj: 71\% BAP \& 63\% DepProbe, obl: 74\% BAP \& 66\% DepProbe. clause: acl: 85\% BAP \& 79\% DepProbe, ccomp: 29\% BAP \& 0\% DepProbe. modifier: advmod: 84\% BAP \& 81\% DepProbe, amod: 77\% BAP \& 66\% DepProbe, discourse: 60\% BAP \& 57\% DepProbe. function: aux: 90\% BAP \& 77\% DepProbe, case: 93\% BAP \& 88\% DepProbe, cop: 97\% BAP \& 71\% DepProbe, det: 93\% BAP \& 93\% DepProbe, mark: 49\% BAP \& 41\% DepProbe. coord: cc: 89\% BAP \& 90\% DepProbe, conj: 66\% BAP \& 43\% DepProbe. multi: compound: 55\% BAP \& 35\% DepProbe, fixed: 66\% BAP \& 56\% DepProbe, flat: 75\% BAP \& 66\% DepProbe. other: punct: 99\% BAP \& 99\% DepProbe, root: 82\% BAP \& 65\% DepProbe.}
    \end{subfigure}
    \hspace{-.7em}
    \begin{subfigure}[m]{.025\textwidth}
        \includegraphics[width=\textwidth]{img/all-targets-correct-rel-legend.pdf}
        \vspace{.2em}
    \end{subfigure}
    \vspace{-1em}
    \caption{\textbf{RelAcc of \bap{} and \depprobe{} on TR-IMST (Test)} grouped according to UD taxonomy.}
    \label{fig:relation-errors-tr}
\end{figure*}

\begin{figure*}
    \begin{subfigure}[m]{.945\textwidth}
        \pdftooltip{\includegraphics[width=\textwidth]{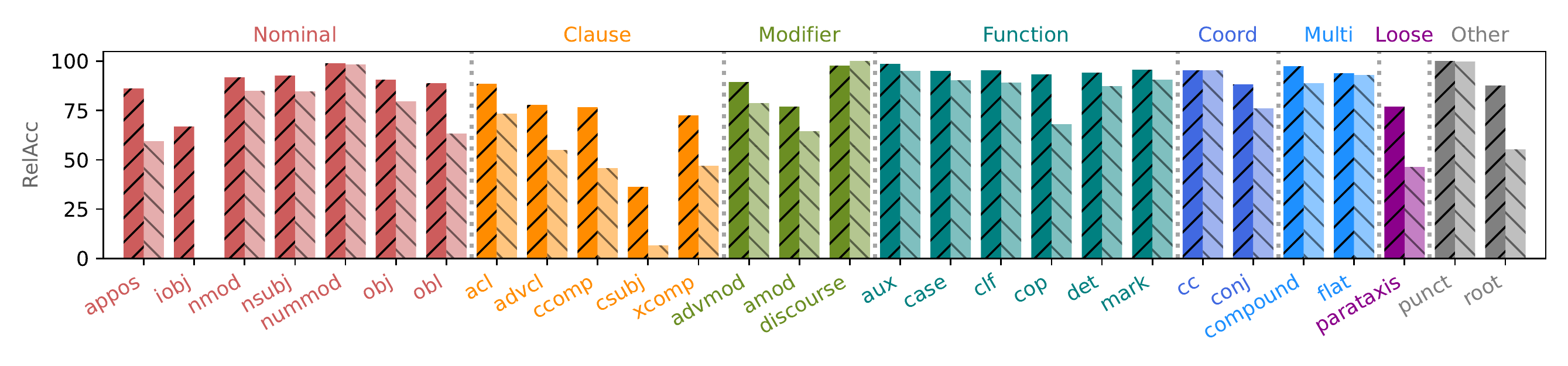}}{Screenreader Caption: nominal: appos: 86\% BAP \& 59\% DepProbe, iobj: 66\% BAP \& 0\% DepProbe, nmod: 91\% BAP \& 84\% DepProbe, nsubj: 92\% BAP \& 84\% DepProbe, nummod: 98\% BAP \& 98\% DepProbe, obj: 90\% BAP \& 79\% DepProbe, obl: 88\% BAP \& 63\% DepProbe. clause: acl: 88\% BAP \& 73\% DepProbe, advcl: 77\% BAP \& 55\% DepProbe, ccomp: 76\% BAP \& 45\% DepProbe, csubj: 36\% BAP \& 6\% DepProbe, xcomp: 72\% BAP \& 46\% DepProbe. modifier: advmod: 89\% BAP \& 78\% DepProbe, amod: 76\% BAP \& 64\% DepProbe, discourse: 97\% BAP \& 100\% DepProbe. function: aux: 98\% BAP \& 94\% DepProbe, case: 94\% BAP \& 90\% DepProbe, clf: 95\% BAP \& 89\% DepProbe, cop: 93\% BAP \& 68\% DepProbe, det: 94\% BAP \& 87\% DepProbe, mark: 95\% BAP \& 90\% DepProbe. coord: cc: 95\% BAP \& 95\% DepProbe, conj: 88\% BAP \& 76\% DepProbe. multi: compound: 97\% BAP \& 88\% DepProbe, flat: 93\% BAP \& 92\% DepProbe. loose: parataxis: 76\% BAP \& 46\% DepProbe. other: punct: 100\% BAP \& 99\% DepProbe, root: 87\% BAP \& 55\% DepProbe.}
    \end{subfigure}
    \hspace{-.7em}
    \begin{subfigure}[m]{.025\textwidth}
        \includegraphics[width=\textwidth]{img/all-targets-correct-rel-legend.pdf}
        \vspace{.2em}
    \end{subfigure}
    \vspace{-1em}
    \caption{\textbf{RelAcc of \bap{} and \depprobe{} on ZH-GSD (Test)} grouped according to UD taxonomy.}
    \label{fig:relation-errors-zh}
\end{figure*}

\paragraph{Relation Accuracy} (RelAcc) is used in Section \ref{sec:dependency-analysis} to analyze dependency relations which benefit from the full parametrization of \bap{} compared to the linear \depprobe{}. Figures \ref{fig:relation-errors-ar}--\ref{fig:relation-errors-zh} show RelAcc per language in addition to the aggregated scores in Figure \ref{fig:relation-errors}. As noted in Section \ref{sec:dependency-analysis}, some relations such as \texttt{cop} differ substantially across languages with respect to their realization (e.g.\ surface form variation). Furthermore, the set of relations represented in each target treebank may differ, especially for specializied categories.

\end{document}